\documentclass[12pt]{article}
\usepackage{amsmath}
\usepackage{times}
\usepackage{graphicx,subcaption}
\usepackage{color}
\usepackage{multirow}
\usepackage{apacite}
\usepackage[authoryear]{natbib}
\usepackage{rotating}
\usepackage{bbm}
\usepackage{latexsym}
\usepackage{epstopdf}
\newcommand{\captionfonts}{\normalsize}

\makeatletter  
\long\def\@makecaption#1#2{%
  \vskip\abovecaptionskip
  \sbox\@tempboxa{{\captionfonts #1: #2}}%
  \ifdim \wd\@tempboxa >\hsize
    {\captionfonts #1: #2\par}
  \else
    \hbox to\hsize{\hfil\box\@tempboxa\hfil}%
  \fi
  \vskip\belowcaptionskip}
\makeatother   

\begin{document}
\hspace{13.9cm}1

\ \vspace{20mm}\\
{\LARGE An Empirical Evaluation of Rule Extraction from Recurrent Neural  Networks}

\ \\
{\bf Qinglong Wang$^{\displaystyle 1}$, 
Kaixuan Zhang$^{\displaystyle 2}$, 
Alexander G. Ororbia II$^{\displaystyle 2}$, \\
Xinyu Xing$^{\displaystyle 2}$, 
Xue Liu$^{\displaystyle 1},$
C. Lee Giles$^{\displaystyle 2}$}
\\
{$^{\displaystyle 1}$McGill University.}\\
{$^{\displaystyle 2}$Pennsylvania State University.}\\

%


\thispagestyle{empty}
\markboth{}{NC instructions}
\ \vspace{-0mm}\\
%
\begin{center} {\bf Abstract} \end{center}
Rule extraction from black-box models is critical in domains that require model validation before implementation, as can be the case in credit scoring and medical diagnosis. Though already a challenging  problem in statistical learning in general, the difficulty is even greater when highly non-linear, recursive models, such as recurrent neural networks (RNNs), are fit to data.  Here, we study the extraction of rules from second order recurrent neural networks trained to recognize the Tomita grammars. We show that production rules can be stably extracted from trained RNNs and that in certain cases the rules outperform the trained RNNs.
\section{Introduction}
Recurrent neural networks (RNNs) have been increasingly adopted for a variety of tasks involving time-varying data, e.g. sentiment analysis, machine translation, image captioning, etc. Despite the impressive performance on these tasks, RNNs are also well-known to be ``black-box'' models, which makes explaining/interpreting the knowledge acquired by these models difficult or near-impossible. This black-box nature is largely due to the fact that RNNs, much as any neural architecture, e.g. convolutional neural networks, although designed to capture structural information from the data~\citep{du2017topology}, store learned knowledge in their weights, which is difficult to inspect, analyze, and verify~\citep{omlin2000symbolic}. 

Given RNN's rising popularity in processing time-varying data, we investigate whether and how we might extract knowledge in symbolic form from RNN models that have been trained on symbolic data, in this case a collection of regular grammars. Surprisingly, this is an old problem treated by Minsky in the chapter titled "Neural Networks. Automata Made up of Parts" in his text "Computation, Finite and Infinite Machines"~\citep{minsky1967computation}. If the information processing procedure of the RNN can be treated as representing knowledge in symbolic form, where a set of rules that govern transitions between symbolic representations are learned, then we can begin to view the RNN as an automated reasoning process that can be easier to understand. Indeed, prior work~\citep{BorgesGL11} has proposed to extract symbolic knowledge from a nonlinear autoregressive models with exogenous (NARX) recurrent model~\citep{Lin96NARX}. For sentiment analysis tasks, recent work~\citep{murdoch2017automatic} has demonstrated that an RNN is capable of identifying consistently important patterns of words. These words can be viewed as symbolic knowledge and the patterns of these words represents the rules for determining the sentiment. In other work~\citep{dhingra2017linguistic}, information about long-term dependencies are also represented in the form of symbolic knowledge to improve the ability of RNNs to handle long-term text data. 
Also, prior work~\citep{giles1992learning, watrous1992induction, omlin1996extraction, casey1996dynamics, jacobsson2005rule} has shown that it is possible to extract deterministic finite automata (DFA) from RNN models trained to perform grammatical inference and that grammatical rules can be stably encoded and represented in second order RNNs~\citep{omlin1996jacm,omlin1996stable}. In these studies, the vector space of an RNN's hidden layer is first partitioned into finite parts, each treated as the states of a certain DFA. Then, transitions rules between these states are extracted. This paper follows the paradigm of DFA extraction laid out in previous research.

While it has been shown that it is possible to extract DFA from RNNs, it has been argued~\citep{kolen1994fool} that DFA extraction is sensitive to the initial conditions of the hidden layer of RNN. In other words, by viewing an RNN as a nonlinear dynamical system, the value of its hidden layer may exhibit exponential divergence for nearby initial state vectors. As a result, any attempts at partitioning the hidden space may result in forcing the extracted state to split into multiple trajectories independent of the future input sequence. This results in an extracted rule that appears as a nondeterministic state transition, even though underlying dynamical system is completely deterministic~\citep{kolen1994fool}.

In this paper, we greatly expand upon previous work in rule extraction from second-order RNNs~\citep{giles1992learning} by studying DFA extraction through comprehensive experiments. The main questions that we hope to ask are:

\begin{enumerate}
	\item What conditions will affect DFA extraction and how sensitive is DFA extraction with respect to these conditions?
    \item How well will the extracted DFA perform in comparison with the RNN trained models from which they are extracted?
\end{enumerate}
\noindent
With respect to the first question, we aim at uncovering the relationship between different conditions, for instance, the influence of the initial condition of the RNN's hidden layer and the configuration of adopted clustering algorithm on DFA extraction. Specially, through our empirical study, we address the concerns of \citep{kolen1994fool} by showing that DFA extraction is very insensitive to the initial conditions of the hidden layer. Moreover, in answering the second question, we find that in most cases the extracted DFA can recognize a set of strings generated by a certain regular grammar as accurately as the trained RNN models from which the DFA were extracted. Interestingly enough, in certain cases, we observe that extracted DFA even outperform their source RNNs in term of recognition accuracy when processing long sequences. This result is surprising given the difficulty in training RNNs on long sequences, largely due to the vanishing gradient problem~\citep{pascanu2013difficulty}, of which a great deal of research has been dedicated to solving \citep{hochreiter1997long,cho2014properties,weston2014memory,sukhbaatar2015end,dhingra2017linguistic}. Extracting rules from RNNs also sheds light on an alternative to improve the processing of long pattern sequences.

Here, our emphasis is on examining the consistency of DFA extraction. More specifically, we first train and test RNN models on data sets generated by the seven Tomita grammars~\citep{tomita1982}. The RNN models we use have a second-order architecture~\citep{giles1992learning}. Then we collect the values of hidden layer units of RNN models obtained during the testing phase, and cluster these values. Here we use k-means due to its simplicity and efficiency. We believe other clustering methods could provide similar results. These clustered states and the symbolic inputs are used to form the initial DFA, which may contain equivalent states. Finally, we use a minimization algorithm to minimize the number of states and finalize the minimal DFA. 

In summary, this work makes the following contributions.
\begin{itemize}
    \item We conduct a careful experimental study of the factors that influence DFA extraction. Our results show that, despite these factors, DFA can be stably extracted from second-order RNNs. In particular, we find strong evidence that, by adopting a simple clustering method, DFA can be reliably extracted even when the target RNN is only trained using short sequences.
    \item We explore the impact of network capacity and training time on the RNN's ability to handle long sequences and find that these factors play key roles. With respect to DFA extraction, however, these factors exhibit only a limited impact. This shows that extracting DFA requires less effort compared to the training of a powerful RNN. 
    \item We investigate a realistic case where ``incorrect'' DFA are extracted from low capacity second-order RNNs, and demonstrate that, in some cases, these DFA can still outperform the source RNNs when processing long sequences. This sheds light on a possible path to improving an RNN's ability in handling long sequences--exploiting the DFA's natural ability to handle infinitely long sequences (which is a challenge for any RNN).
\end{itemize}
\section{Background}
\label{sec:background}
Recurrent neural networks (RNN) process sequential data by encoding information into the continuous hidden space in an implicit, holistic manner~\citep{elman1990finding}. In order to extract rules from this continuous hidden space, it is commonly assumed that the continuous space is approximated by a finite set of states~\citep{jacobsson2005rule}. The rule is then referred to as the transitions among the discrete states. A common choice for representation of the extracted rules is a DFA. In the following, we first provide a brief introduction of DFA, followed by an introduction to the target grammars studied. Finally, we present a particular type of RNN -- a second-order RNN, which is mainly used in this work. 

\subsection{Deterministic Finite Automata}
A finite state machine $M$ recognizes and generates certain grammar $G$, which can be described by a five-tuple $\{A, S, s_0, F, P\}$. Here, $A$ is the input alphabet (a finite, non-empty set of symbols), $S$ is a finite, non-empty set of states. $s_0 \in S$ and $F \in S$ represents the initial state (an element of $S$) and the set of final states (a subset of $S$, $F$ can be empty). $P$ denotes a set of production rules (transition function $P:S\times A \rightarrow S$). Every grammar $G$ also recognizes and generates a corresponding language $L(G)$, a set of strings of the symbols from alphabet $A$. The simplest automata and its associated grammar are DFA and regular grammars, according to the Chomsky hierarchy of phrase structured grammars~\citep{chomsky1956three}. It is important to realize DFA actually covers a wide range of languages, that is, all languages whose string length and alphabet size are bounded can be recognized and generated by finite state automata~\citep{giles1992learning}. Also, when replacing the deterministic transition with stochastic transition, a DFA can be converted as a probabilistic automata or hidden Markov model, which enables grammatical inference as learning on graphical models~\citep{du2016convergence}. We refer the reader to a more detailed introduction of regular languages and finite state machines in ~\citep{hopcroft2006automata,carroll1989theory} and use their notation.

\subsection{Tomita Grammars}
We select a set of seven relatively simple grammars, which are originally suggested by Tomita~\citep{tomita1982} and widely studied~\citep{pollack1991induction,omlin1996extraction,watrous1992induction} and use them for an empirical study for extracting rules from RNN. We hypothesize (and note from the work of others) that these simple regular grammars should be learnable. More specifically, the DFA associated with these grammars have between three and six states. These grammars all have $A = \{0,1\}$, and generate an infinite language over $\{0,1\}^{*}$. Here we denote a finite set of strings $I$ from regular language $L(G)$. Positive examples of the input strings are denoted as $I_{+}$ and negative examples as $I_{-}$. We provide a description of positive examples accepted by all seven grammars in Table~\ref{tab:tomita}.
\begin{table}[t]
\centering
\caption{Description of seven Tomita grammars}
\label{tab:tomita}
\small
\begin{tabular}{ll}
\hline \hline
G & Description                                                                              \\ \hline \hline
1 & $1^{*}$                                                                                  \\ \hline
2 & $(1 0)^{*}$                                                                              \\ \hline
3 & \begin{tabular}[c]{@{}l@{}}an odd number of consecutive 1s is always followed by \\ an even number of consecutive 0s\end{tabular} \\ \hline
4 & any string not containing ``000'' as a substring                                         \\ \hline
5 & even number of 0s and even number of 1s~\citep{giles1990higher}                          \\ \hline
6 & the difference between the number of 0s and the number of 1s is a multiple of 3          \\ \hline
7 & $0^{*}1^{*}0^{*}1^{*}$                                                                   \\ \hline \hline
\end{tabular}
\end{table}

The associated DFA for these grammars is shown in the first column in Figure~\ref{fig:visual_dfa}. Some of these DFA contain a so-called ``garbage state'', that is, a non-final state in which all transition paths lead back to itself. In order to correctly learn this state, RNN must not only learn with positive strings $I_{+}$ generated by the grammar, but also negative strings $I_{-}$ that are rejected by this grammar. 

Despite the fact that the Tomita grammars are relatively simple, we select these grammars because they actually cover regular languages that have different complexity and difficulty~\citep{wang2018model}. They also appear to be a standard for much work on learning grammars. For example, grammars 1,2 and 7 represent the class of regular languages that define a string set that has extremely unbalanced positive and negative strings. This implies that the averaged difference between positive strings and negative strings can be very large. This could represent real-world cases where positive samples are significantly outnumbered by negative ones. In contrast, grammars 5 and 6 define the class of regular languages that have equal or a relatively balanced number of positive and negative strings. This implies that the difference between positive and negative strings in these grammars is much smaller than the case of grammars 1,2 and 7. Finally, grammars 3 and 4 represent the class of regular languages for which the difference between positive and negative strings is somewhere between the above two cases. With either case discussed, the source RNNs are forced to recognize the various levels of difference between positive and negative samples. In addition, it is also important to note that we have ground truth DFAs for Tomita grammars. This enables this study to determine the impact of different factors on the success rate of extracting correct DFAs (introduced in Section 4), since they can be compared to the ground truth DFAs. With more complex/or real-world datasets, this may not be case. For those datasets, uncertainties will be introduced into the evaluation (e.g. what is the ground truth DFA or if there even exists ground truth DFAs that define the data?). This uncertainty can affect any conclusion of whether a DFA extraction can be stably performed. 

\subsection{Second-order Recurrent Neural Networks}
Here, we use an RNN constructed with second-order interactions between hidden states and input symbols. More specifically, this second-order RNN has a hidden layer $H$ containing $N$ recurrent hidden neurons $h_i$, and $L$ input neurons $i_l$. The second-order interaction is represented as $w_{ijk} h_{j}^{t} i_{k}^{t}$, where $w_{ijk}$ is a $N\times N\times L$ real-valued matrix, which modifies a product of the hidden $h_j$ and input $i_k$ neurons. $t$ denotes the $t$th discrete time slot. This quadratic form directly represents the state transition diagrams of a state process -- $\{$input, state$\} \Rightarrow \{$next state$\}$. More formally, the state transition is defined by following equation:
\begin{equation}
H_{i}^{t+1} = g(\sum_{j,k} W_{ijk} H_{j}^{t} I_{k}^{t})
\end{equation}
where $g$ is a sigmoid discriminant function. Each input string is encoded by one-hot-encoding, and the neural network is constructed with one input neuron for each character in the alphabet of the relevant language. By using one-hot-encoding, we ensure that only one input neuron is activated per discrete time step $t$. Note that, when building a second-order RNN, as long as $L$ is small compared to $N$, the complexity of the network only grows as $O(N^2)$. Such RNNs have been proved to stably encode finite state machines~\citep{omlin1996jacm,omlin1996stable} and thus can represent in theory all regular grammars.

To train above second-order RNN, we use the following loss function $C$ following~\citep{giles1992learning}:
\begin{equation}
C = \frac{1}{2}(y - H_{0}^{T})^{2}
\end{equation}
where $C$ is defined by selecting a special ``response'' neuron $h_{0}$, which is compared to the target label $y$. For positive strings, $y = 1.0$ and $y = 0.0$ for negative strings. $h_{0}^{T}$ indicates the value of $h_0$ at time $T$ after seeing the final input symbol. We adopt \textit{RMSprop}~\citep{tieleman2012lecture} as the training algorithm. 
\section{DFA Extraction}
\label{sec:method}
We introduce our approach to DFA extraction, which largely builds on the research conducted in the 1990's (please see many of the citations in the bibliography) but note that there has been recent work~\citep{principe2016}. We start by briefly introducing the main ideas behind DFA extraction as well as existing research. We will then examine and identity key factors that affect the quality of each step of the extraction process.

\subsection{The DFA Extraction Paradigm}
\label{sec:paradigm}
Many methods have been developed to extract knowledge in the form of rules from trained RNNs~\citep{giles1991second,giles1992learning,omlin1996extraction,zeng1993learning,frasconi1996representation,gori1998inductive}. Most of this work can be viewed as roughly following one general DFA extraction process:
\begin{enumerate}
	\item Collect the hidden activations of the RNN when processing every string at every time step.  Cluster these hidden activations into different states.
    \item Use the clustered states and the alphabet-labeled arcs that connect these states to construct a transition diagram.
    \item Reduce the diagram to a minimal representation of state transitions.
\end{enumerate}
Previous research has largely focused on improving the first two steps. This is largely due to the fact that for the third step, there already exists a well established minimization algorithm~\citep{hopcroft2006automata} for obtaining the minimal representation of a DFA.

For the first step, an equipartition-based approach~\citep{giles1992learning} was proposed to cluster the hidden space by quantizing the value of a hidden unit to a specified number of bins. For example, if we apply a binary quantization~\footnote{Using a threshold value of $0.5$, any value greater than $0.5$ is assigned to the bin ``1'', whereas other values less than or equal to this threshold are assigned to ``0''.} to the vector $\{0.6, 0.4, 0.2\}$, we would obtain the encoding $\{1, 0, 0\}$. One drawback of this form of quantization is that as the number of hidden units increases, the number of clusters grows exponentially. This computational complexity issue is alleviated if one uses clustering methods that are less sensitive to the dimensionality of data samples, e.g., k-means~\citep{zeng1993learning, frasconi1996representation, gori1998inductive}, hierarchical clustering~\citep{sanfeliu1994active}, and self-organizing maps~\citep{tivno1995learning}.

In order to construct state transitions for the second step, either breadth-first search (BFS) approaches \citep{giles1992learning} or sampling-based approaches~\citep{tivno1995learning} are utilized. BFS approaches can construct a transition table that is relatively consistent but incur high computation cost especially when the size
of alphabet increases exponentially. Compared with the BFS approach, the sampling approach is computationally more efficient. However, it introduces inconsistencies in the construction of a transition table. For a more detailed exposition of these two classes of methods, we refer the readers to this survey~\citep{jacobsson2005rule}. 

\subsection{Factors That Affect DFA Extraction}
\label{sec:factors}

The efficacy of the different methods used for the first two steps of the process described above rely on the following hypothesis: The state space of a well-trained RNN should already be fairly well-separated, with distinct regions or clusters that represent corresponding states in some DFA. This hypothesis, if true, would greatly ease the process of DFA-extraction. In particular, less effort would be required in the first two steps of DFA extraction if the underlying RNN was constructed to have a well-separated state-space. 

With this in mind, we specify the following key factors that affect DFA extraction that also affect representational ability of an RNN.
\begin{itemize}
	\item Model capacity. A RNN with greater capacity (larger size of hidden layer) is more likely to better represent a DFA.
    \item Training time. A sufficient number of iterations are required in order to ensure convergence (to some local optima).
    \item Initial conditions of the hidden state. As argued previously~\citep{kolen1994fool}, the initial conditions may have significant impact on DFA extraction. In this work, we explore this impact by training several RNN models with random initial hidden activations on all grammars, then examining the extracted DFA from all trained RNN models.
    \item Choice of state-clustering method. The choice of clustering algorithm is very important, including its hyper-parameter configuration. For example, if k-means or a Gaussian mixture model is adopted, a critical hyper-parameter is the predefined number of clusters.
\end{itemize}
One could argue that other factors, such as choice of a parameter update rule (e.g., ADAM, RMSProp, etc.) and learning rate, may also influence how well an RNN learns about certain grammar. However, in our experiments, we observe that these latter conditions actually have little and nearly no influence on the final results. Thus, we focus on the factors described in the list above.

\subsection{The DFA Extraction Process}
\label{sec:our_approach}
Here, we use an approach similar to~\citet{zeng1993learning} to extract DFA from second-order RNNs. To be more specific, we first train second-order RNNs to classify strings generated by each of the seven Tomita grammars \citep{tomita1992dynamic}. A desirable outcome of the hypothesis described in the previous section is that, when the hidden space is well-separated, many well-established clustering methods should generate similar results. This allows us to choose our clustering approach based on computational complexity. As a result, we adopt the k-means clustering approach due to its simplicity and efficiency. However, as will be discussed in the following, we must now concern ourselves with choosing an appropriate value of $K$.

After clustering the hidden space, we follow the approach taken in \citet{schellhammer1998knowledge} to construct the transition diagram. Specifically, we construct the diagram by counting the number of transitions that have occurred between a state and its subsequent states (given a certain input). For example, given a state $S_k$ and input symbol $i$, we calculate the number of transitions to all states $\{S\}$ from $S_k$, including any self-loops. After obtaining the transition counts, we keep only the most frequent transitions between $\{S\}$ and $\{S+1\}$ given input $i$ and discard the other less frequent ones in the transition diagram. 

\begin{figure}[t!]
\hfill
\begin{center}
\includegraphics[width=1.0\linewidth]{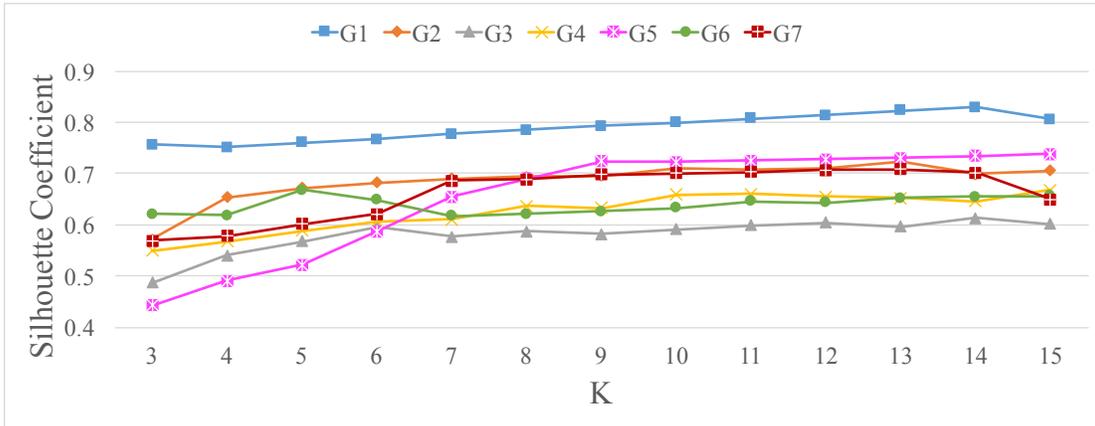}
\end{center}
\caption{Influence of $K$ on clustering results for all grammars}
\label{fig:silhouette}
\end{figure}

It is important to note that $K$ should not be set too small. In an extreme case, when the value of $K$ is set to be even smaller than the minimal number of states of the ground truth DFA, the extraction never provides the correct DFA. Additionally, when $K$ is small, the hidden activations that should have formed different clusters (which represent different states) may be forced to be included in a single cluster, hence generating poor clustering. We illustrate this effect by demonstrating in Figure~\ref{fig:silhouette} the clustering obtained by selecting different $K$'s. More specifically, we evaluate the clustering using a silhouette coefficient to measure how well the resulting clusters are separated. As shown in Figure~\ref{fig:silhouette}, when $K$ is smaller than 6, the clustering is much less desirable and varies significantly than when $K$ is larger. This poorly clustered hidden space will more likely cause inconsistent transitions between states given the same input. For example, assuming there are two cluster $S_1$ and $S_2$, given the same input symbol $i$, they transit to $S_3$ and $S_4$ respectively. When $K$ is small, it is possible that $S_1$ and $S_2$ are merged as one cluster $\hat{S}_1$. As a result, $\hat{S}_1$ will inconsistently visit $S_3$ and $S_4$ with the same input $i$. This falsely indicates that the transition learned is more likely to be non-deterministic, while the real case is that the RNN generates $S_{t+1}$ based on $S_{t}$ and $i$ deterministically. This effect can be mitigated when $K$ is increased beyond a certain value. However, this does not indicate that $K$ can be set arbitrarily large. Larger $K$ only brings limited improvement in the clustering results, while imposing more computation on both the clustering algorithm and the minimization algorithm which is introduced next.

With the constructed transition diagram, we have extracted a DFA that might contain many redundant states. Using the previously described minimization algorithm~\citep{hopcroft2006automata}, we can then reduce the derived DFA to its minimal representation. Note that this minimization algorithm does not change the performance of the DFA; the unminimized DFA has the same time complexity as the minimized DFA. Note that the DFA extraction method introduced above may be applied to any RNN, regardless of order or manner in which its hidden layers are calculated.

\section{Experiments}
\label{sec:eval}
In this section, we empirically study the process of DFA extraction through comprehensive experiments.

\subsection{Description of Data}
\label{sec:data}

To train and test the RNN models, we followed the approach introduced in~\cite{giles1992learning} and generated string sets. To be specific, we drew strings from an oracle generating random 0 and 1 strings and the grammar specified in Table~\ref{tab:tomita}. The end of each string is set to symbol 2, which represents the ``stop'' symbol (or end-token as in language modeling). For the strings drawn from a grammar, we took them as positive samples while those from that random oracle as negative samples. Note that we verified each string from the random oracle and ensured they are not in the string set represented by that corresponding grammar before treating them as negative samples. It should be noticed that for each grammar in our experiments represents one set of strings with unbounded size. As such we restricted the length of the strings used with a upper bound equal to 15 for all grammars. In addition, we also specify a lower bound on the string lengths to avoid training RNNs with empty strings. In order to use as many strings as possible to build the datasets, the lower bound should be set to be sufficiently small. In our experiments, we set the lower bound equal to 3 for all the grammars. We split the strings generated within the specified range of length for each grammar to build the training set $D_{train}$ and testing set $D_{test}$, then trained and tested the RNNs accordingly.

In order to further the trained RNNs and extracted DFA on longer strings, we build another testing set $D_{test(200)}$ comprised of strings of length 200 for all grammars. Note that the complete set of strings with length 200 numbers around $10^{60}$. A test set of this size is too expensive and not even necessary for evaluating RNNs or DFA. Therefore, we construct the testing set by randomly sampling 100,000 strings for all grammars. In addition, to preserve the actual balance of positive to negative samples, we sample such that we preserve their original proportions as measured from the original, complete set of length 200 strings. For example, for grammar 5, we sample positive and negative strings with the same ratio of 0.5. 

\subsection{The Influence of Model Capacity}
\label{sec:model_complexity} 
\begin{figure}[t]
\centering
\begin{subfigure}{1.0\textwidth}
  \centering
  \includegraphics[width=1.0\linewidth]{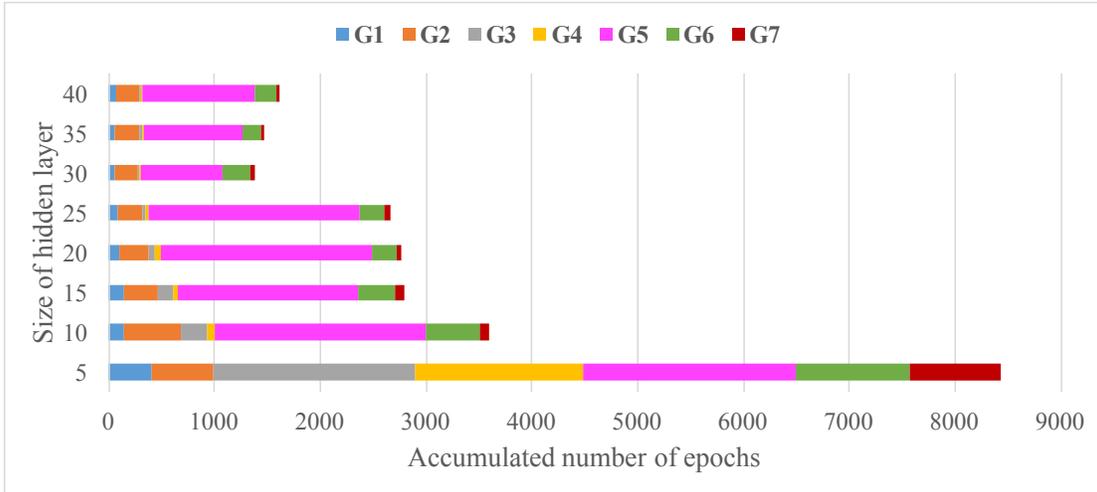}
  \caption{Influence on training RNN.}
  \label{fig:capacity_k}
\end{subfigure} \\%
\begin{subfigure}{1.0\textwidth}
  \centering
  \includegraphics[width=1.0\linewidth]{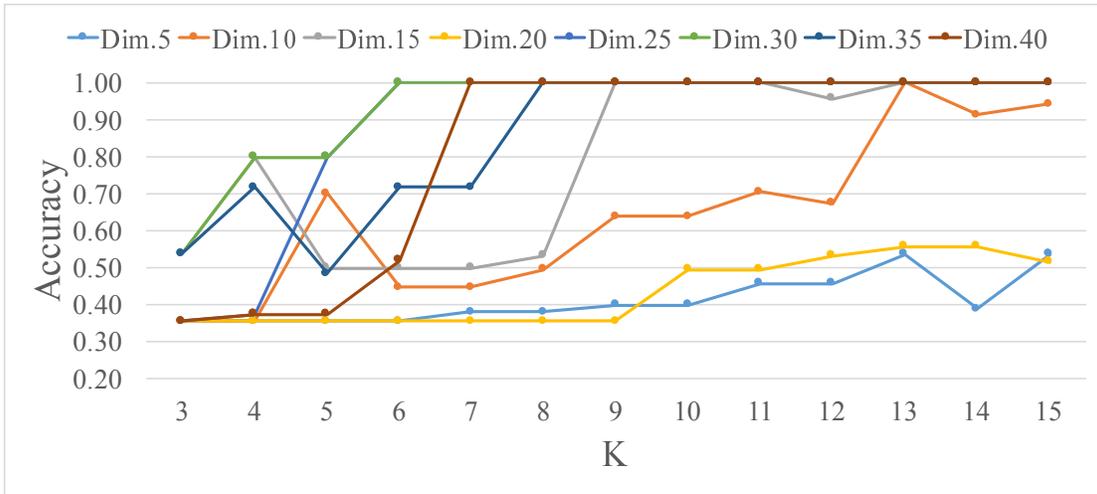}
  \caption{Influence on extracting DFA.}
  \label{fig:g3_capacity_k}
\end{subfigure}
\caption{The influence of model capacity on DFA extraction for grammar 3.}
\label{fig:dfa_complexity}
\end{figure}

In following experiment, we first measure the influence of model capacity, i.e. the size $N$ of hidden layer of RNN models, on learning the target DFA. Specifically, we measure the training time needed for RNNs with different hidden layer sizes to reach perfect accuracy on the testing set $D_{test}$ for all grammars. It is clear from Figure~\ref{fig:capacity_k} that it takes less training for an RNN with a larger capacity $N$ to converge. This is what we would expect; an RNN with larger capacity, in general, can better fit the data.

Next, we evaluate how stably correct DFA can be extracted from the trained RNN models. Here we argue that DFA extraction should be more stable from a RNN model for which the hidden state space is well separated. Intuitively, a well-separated space means that with a well trained second-order RNN, hidden activations obtained after processing a set of strings have already aggregated into different regions that are separated from each other in the hidden node space. In this case, it would be more flexible to select a different $K$ to cluster this space. Assuming the ground truth value of $K$ is $M$, as when $K$ is larger than $M$, K-means can already identify $M$ large clusters that contain the majority of the hidden activations. For the other $M - K$ clusters, they only identify outliers. This is also verified in Figure~\ref{fig:silhouette}. Specifically, when $K$ is sufficiently large, the silhouette coefficient changes slightly as $K$ increases. This is because the small clusters formed by outliers only contribute trivially to the calculated silhouette coefficient. These small clusters will later be recognized as a redundant state and are eliminated by the minimization algorithm. As such, we believe that a more flexible choice of $K$ indicates that the hidden space has already been well separated. To examine this flexibility, we vary $K$ within a certain range to check the accuracy of the extracted DFAs. When more correct DFAs can be extracted from a model, we then determine the choice of $K$ as being more flexible, thus indicating that this model has its hidden space better separated. 

From the above discussion for models with a different number of hidden neurons, we compare the classification accuracy on $D_{test}$ of the extracted DFA when increasing $K$ from 6 to 15 on grammar 3. Similar results for the other grammars are provided in the Appendix. As shown in Figure~\ref{fig:g3_capacity_k}, models with the number of hidden neurons larger than 10 allow more flexible choices for $K$. For instance, when $N>20$, the correct DFA can be reliably extracted in most cases of $K$ from 3 to 16. On the contrary, for models with less hidden neurons, the range of $K$ that produces correct DFAs is more limited. For instance, when $N = 5$, the extraction fails for all $K$ within the same range. In addition, when $N$ is larger than 25, successful extraction is only observed when $K$ is larger than 8. These results also indicate that DFA extraction is more likely to succeed when $K$ is set to larger values. This observation is consistent with the results reported in~\citet{zeng1993learning}. 

The above experimental results indicate that RNNs with larger capacity are more likely to automatically form a reasonably well-separated state space. As a result, the extraction of DFA is less sensitive to the hidden state clustering step of the process. 

\subsection{The Influence of Training Time}
\label{sec:train_time}

\begin{table}[t!]
\small
\centering
\caption{Influence of training time on DFA extraction and RNN performance.}
\label{tab:perf_train_time}
\small
\begin{tabular}{ccccccccc}
\hline \hline
Grammar             & \multicolumn{8}{c}{Classification errors reached under different training epochs.} \\ \hline \hline
\multirow{4}{*}{G3} & Epoch     & 10        & 20        & 30       & 40       & 50     & 60     & 70     \\ \cline{2-9} 
                    & RNN($D_{test}$)       & 0.99      & 0.99        & 0.99       & 1.0     & 1.0     & 1.0      & 1.0      \\ \cline{2-9} 
                    & RNN($D_{test(200)}$)       & 9.1e-2      & 0.55        & 0.51       & 0.81     & 1.0     & 0.96      & 0.86      \\ \cline{2-9} 
                    & DFA       & 1.0     & 1.0         & 1.0        & 1.0        & 1.0      & 1.0      & 1.0      \\ \hline \hline
\multirow{4}{*}{G4} & Epoch     & 10         & 20        & 30       & 40       & 50     & 60     & 70     \\ \cline{2-9} 
                    & RNN($D_{test}$)       & 0.98     & 0.99     & 0.99        & 1.0      & 1.0      & 1.0      & 1.0      \\ \cline{2-9} 
                    & RNN($D_{test(200)}$)       & 3.3e-3     & 3.6e-3     & 4.4e-3        & 9.7e-3      & 4.1e-3      & 8.9e-3      & 7.8e-3      \\ \cline{2-9} 
                    & DFA       & 1.4e-3     & 1.2e-3         & 2.4e-3        & 1.0        & 1.0      & 1.0      & 1.0      \\ \hline \hline
\multirow{4}{*}{G5} & Epoch     & 600       & 650       & 700      & 750      & 800    & 850    & 900    \\ \cline{2-9} 
                    & RNN($D_{test}$)       & 0.63      & 0.29      & 0.46     & 0.44     & 1.0      & 1.0      & 1.0      \\ \cline{2-9} 
                    & RNN($D_{test(200)}$)       & 0.63      & 0.3      & 0.46     & 0.45     & 1.0      & 1.0      & 1.0      \\ \cline{2-9} 
                    & DFA       & 0.67      & 0.67      & 0.67     & 0.67     & 1.0      & 1.0      & 1.0      \\ \hline \hline
\end{tabular}
\end{table}

In this part, we evaluate the classification performance of both trained RNNs and extracted DFA when processing longer strings. More specifically, we measure the classification errors made by both RNNs and DFA on the test set $D_{test(200)}$, as shown in Table~\ref{tab:perf_train_time}. For example, with respect to grammar 3, we train seven RNNs with different training epochs (increasing from 10 to 70). Seven DFA are then extracted, the testing performance of each as a function of epoch is displayed in Table~\ref{tab:perf_train_time}. Due to the space restriction, here we only show the results obtained for grammars 3,4 and 5. The results for other grammars are provided in the Appendix.

As expected, as the training time increases, RNNs tend to make more accurate classification. In particular, for grammars 3, 4 and 5, the trained RNNs reach 100\% accuracy on $D_{test}$. We observe that the correct DFA can sometimes be extracted even when the RNN has not yet fully reached 100\% accuracy (10th to 30th epoch for grammar 3). This indicates that the hidden state space learned in the early stages of training (before the RNN is fully trained) can still be sufficient for a clustering method to recognize each individual state. This observation implies that less effort is needed to extract the correct DFA from an ``imperfect'' RNN than in training a ``perfect'' RNN.

Two other interesting observations can be made with respect to grammar 5. First, it takes much longer to both train an RNN to achieve perfect classification performance and extract the correct DFA from it. Second, the correct DFA can only be successfully extracted when the source RNN quits making any mistakes on the test sets. The difficulty behind training on grammar 5 might be explained through examination of the ``differences'' between the positive and negative strings generated by the grammar. More specifically, by flipping 1 bit of a string from 0 to 1, or vice versa, any positive (negative) string can be converted to a negative (positive) string. In order to learn the small differences, an RNN needs significantly more training time. The second observation may be explained by noting that, before reaching 800 epochs, RNNs make a nearly constant number of errors. This clearly indicates that the RNN is stuck at a certain local minima (also verified in Figure~\ref{fig:train_loss}). While the training of RNN is trapped in this minima, the state space does not start to form the correct partition. However, after 800 epochs, the model escapes this minima and finally converges to a better one, resulting in a state space that is separated correctly.

\subsection{The Influence of Initial States \& Clustering Configuration}
\label{sec:random_init}

In the following experiments, we examine if a DFA can be stably extracted under random initial conditions. Specifically, for each grammar, we randomly assign an initial value to the hidden activations, i.e. $H_{0:N}^{0}$ at $t_0$ time step, within the interval of $[0.0,1.0]$. We repeat this random initialization ten times (training ten different RNNs) for each grammar. Furthermore, we vary the value of $K$ for the k-means clustering algorithm, measuring the classification performance of each extracted DFA, and counting the number of times the correct DFA is extracted (only DFA achieving 100\% accuracy are regarded as correct). Through this procedure, we hope to uncover the relationship between the initial condition of the RNN's hidden layer as well as the clustering algorithm's meta-parameter $K$ and DFA extraction. 

\begin{figure}[!t]
\centering
\begin{subfigure}{.80\textwidth}
  \centering
  \includegraphics[width=\linewidth]{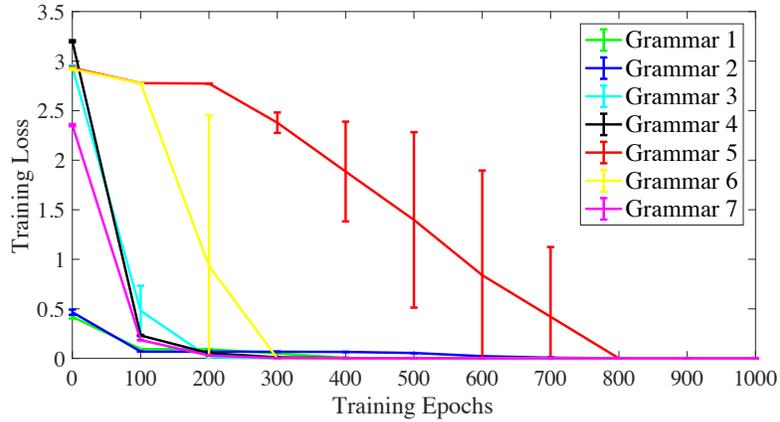}
  \centering
  \caption{Mean and variance of training loss of RNNs on all grammars.}
  \label{fig:train_loss}
\end{subfigure} \hfill
\begin{subfigure}{.80\textwidth}
  \centering
  \includegraphics[width=\linewidth]{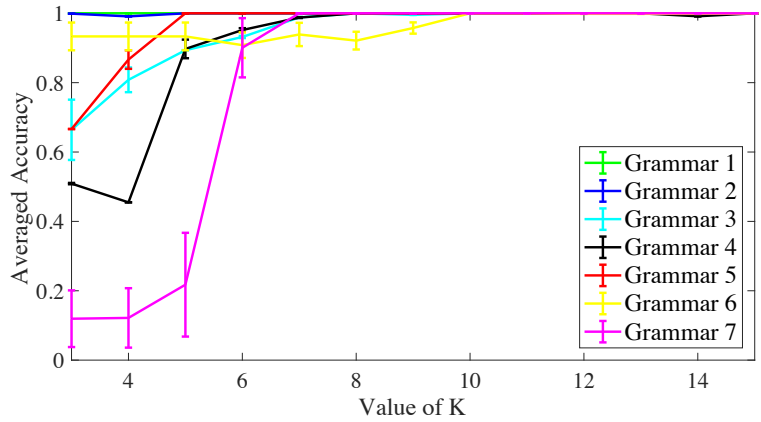}
  \caption{Mean and variance of testing accuracy of extracted DFA with varying K on all grammars.}
  \label{fig:extract_success}
\end{subfigure} \hfill 
\caption{Influence of random initialized hidden activations and clustering configuration on training RNN and extracting DFA.}
\label{fig:avg_var}
\end{figure}

As previously discussed, training an RNN properly is critical for successful DFA extraction. In Figure~\ref{fig:train_loss}, we show the mean and variance of the training loss obtained when training each RNN with 10 times of random initialization of hidden activation for all grammars. It is clear from Figure~\ref{fig:train_loss} that, except for grammar 5 and 6, RNNs trained on other grammars rapidly converge. For grammar 5 and 6, RNNs need much more training time while having much larger variance of training loss. Recall above discussion in Section~\ref{sec:train_time}, this is a clearer indication that the training of these RNNs is trapped to different local optima with different initial activation. However, when given sufficient training, all RNNs trained on all grammars converge on the training set and reach 100\% accuracy on $D_{test}$. In addition, once these RNNs converge to small training loss, the variance reduce to almost 0. This indicates that, with sufficient training and reasonable capacity, RNN's training is relatively insensitive to the hidden layer's initialization.

Given the RNNs trained as described above, we then vary $K$ as we extract DFA from these models. Similarly, we report the mean and variance of the classification accuracy obtained on $D_{test}$ from all extracted DFA in Figure~\ref{fig:extract_success}. To be more specific, for each grammar, under each random initialization of the model's hidden layer, we run the extraction process 13 times, varying $K$ in the range from 3 to 15. In total, we conduct 130 rounds of DFA extraction from the ten trained RNNs for each grammar.

As shown in Figure~\ref{fig:extract_success}, when $K$ is set to small values (below 8), except for grammar 1 and 2, the extracted DFA on other grammars have not only poor classification accuracy, but also relatively large variance. In this case, it is difficult to determine whether random initialization of hidden activation or $K$ have stronger impact on the extraction. When $K$ is set to a sufficiently large value, however, the variance is significantly reduced while the classification accuracy is greatly improved. This indicates that a sufficiently large $K$ can offset the impact of initial states. 
\begin{figure}[!t]
\hfill
\begin{center}
\includegraphics[width=1.0\linewidth]{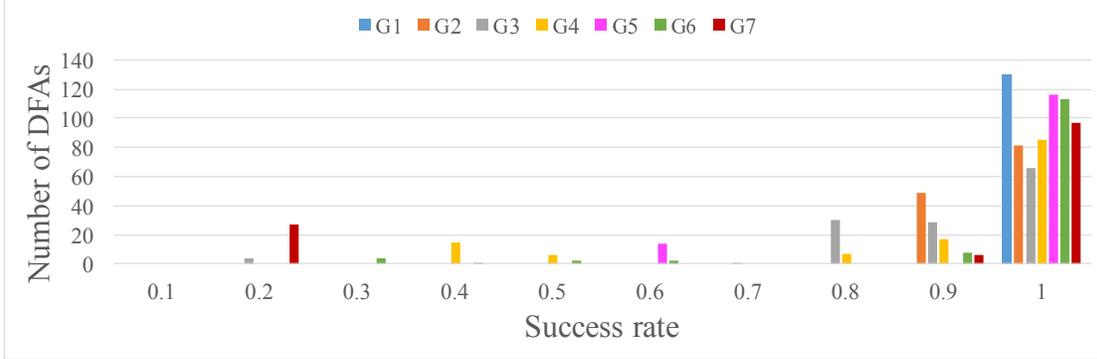}
\end{center}
\vspace{-1.6em}
\caption{Histograms of the classification performance of extracted DFA on all grammars. The vertical axis is displayed in log-scale.}
\label{fig:success_bar}
\end{figure}

\begin{figure}[!t]
\hfill
\begin{center}
\includegraphics[width=5.0in]{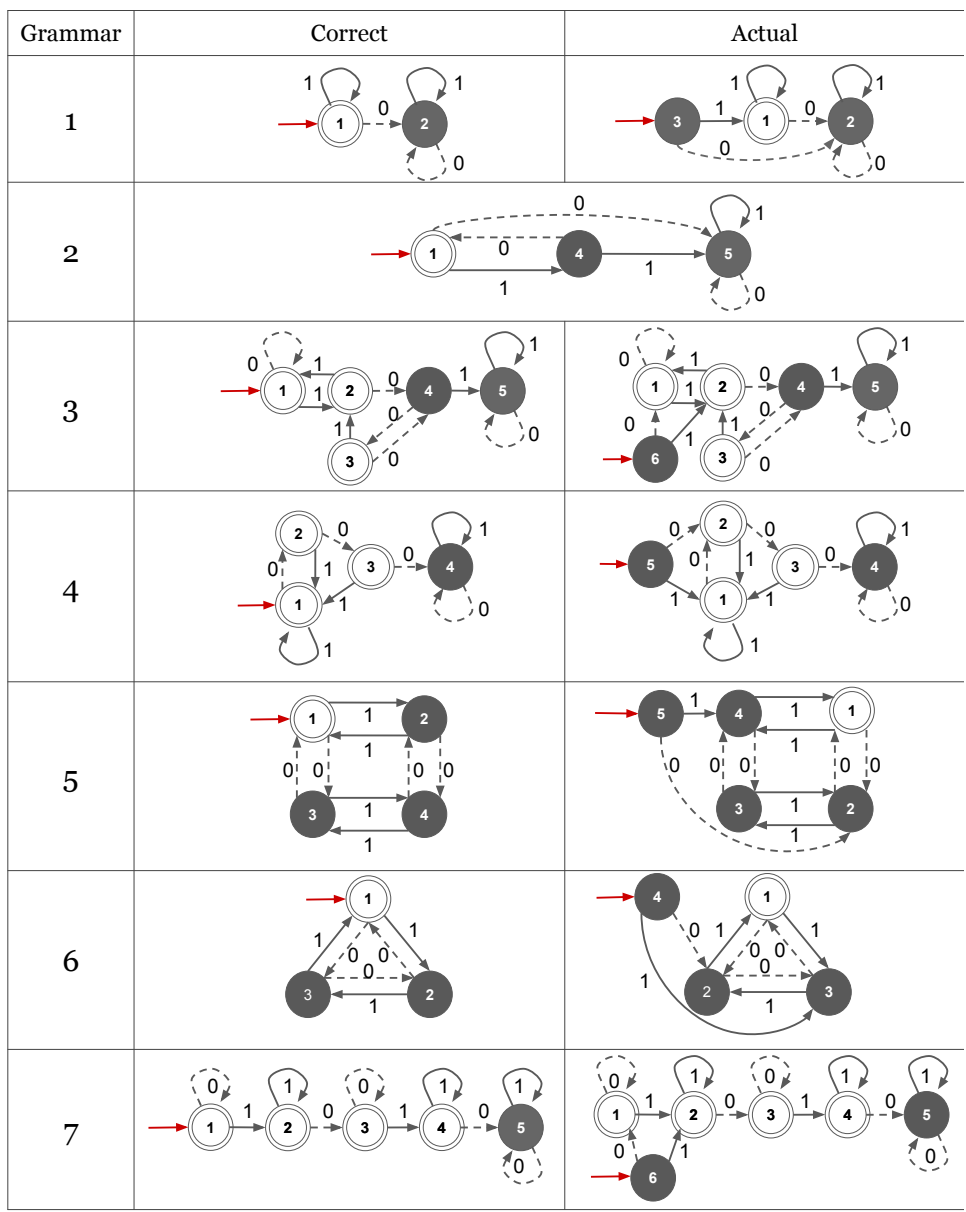}
\end{center}
\caption{Visualization of ground truth DFA and extracted DFA for all grammars. Dotted lines indicate input symbol ``0'' and solid lines indicate input symbol ``1''.}
\label{fig:visual_dfa}
\end{figure}

Besides showing the classification performance obtained by the extracted DFA, we further measure the success rate of extraction in Figure~\ref{fig:success_bar} under different $K$. More specifically, the success rate of extraction is the percentage of DFAs with 100.0\% accuracy among all DFAs extracted for each grammar under different settings of $K$ and random initializations. Among all 130 rounds of extraction on each grammar, we observe that the correct DFA is successfully extracted with highest success rate of 100.0\% (on grammar 1), lowest success rate of 50.0\% (on grammar 3) and averaged success rate of 75.0\% among all grammars. The reason for the worse extraction results obtained on grammar 3 can be explained by visualizing the extracted DFA in Figure~\ref{fig:visual_dfa}. 

In Figure~\ref{fig:visual_dfa}, we see that all extracted DFA can correctly recognize their associated positive strings, except for ones have length smaller than the minimal length we set. Recall that for all grammars, we generate strings while constraining the minimal string length. The visualization indicates that the extracted DFA not only accurately represent the target grammar that generated the string samples, but also obey the constraint on length. In order for the extracted DFA to satisfy the minimal length constraint, extra states are required, as shown in right panel of Figure~\ref{fig:visual_dfa}. Especially, for grammar 3, 4 and 7, the correct DFA contain 5, 4 and 5 states, while the corresponding extracted DFA have 6, 5 and 7 states, respectively. Recall that in above experiments, the minimal value of $K$ is set to 3 consistently for all grammars. As a result, this setting of $K$ causes many failures of extraction for these grammars. As shown in Figure~\ref{fig:extract_success}, when $K$ is below 8, the averaged classification accuracy of the extracted DFA are relatively lower in comparison with DFA extracted from other grammars.

\subsection{Comparison between Low-Capacity RNNs and Extracted DFA}
\label{sec:compare_rnn_dfa}

\begin{figure}[t]
\begin{subfigure}{.50\textwidth}
  \centering
  \includegraphics[width=0.9\linewidth]{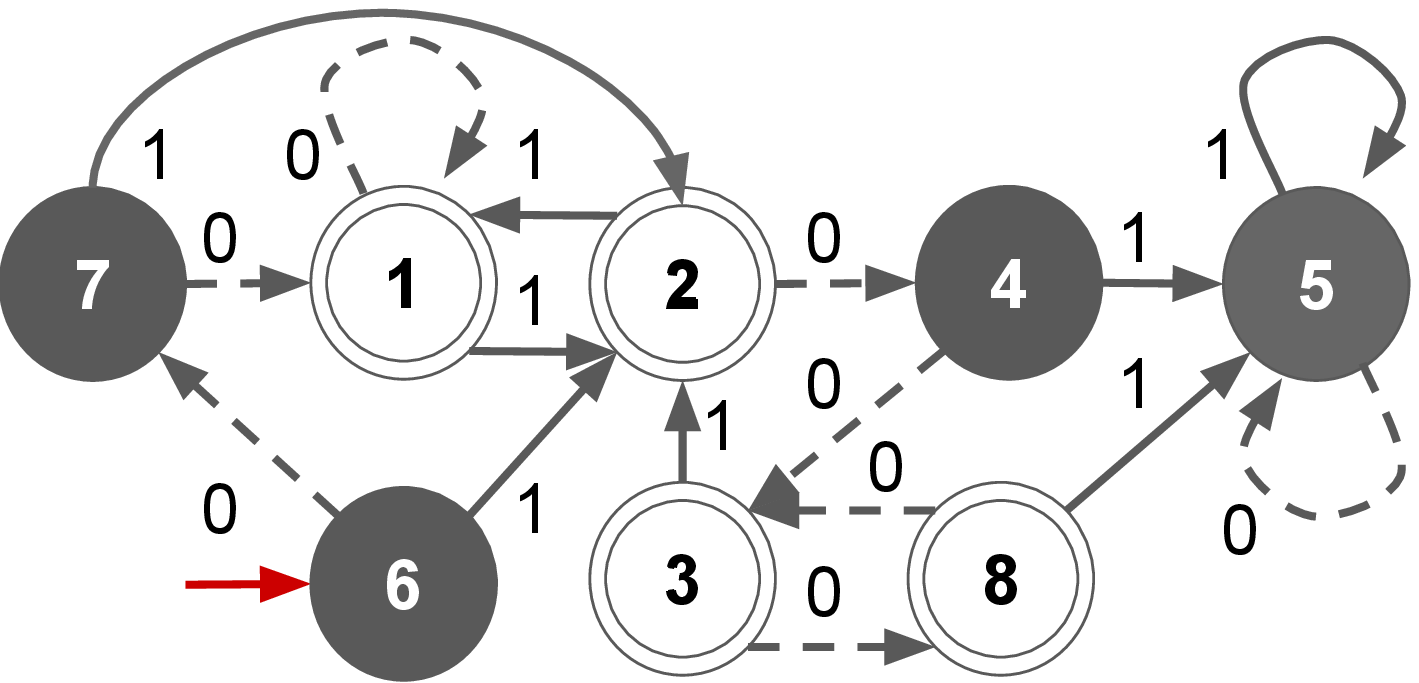}
  \caption{Incorrect DFA extracted on grammar 3.}
  \label{fig:Wrong_DFA_g3}
\end{subfigure} \hfill%
\begin{subfigure}{.45\textwidth}
  \centering
  \includegraphics[width=0.9\linewidth]{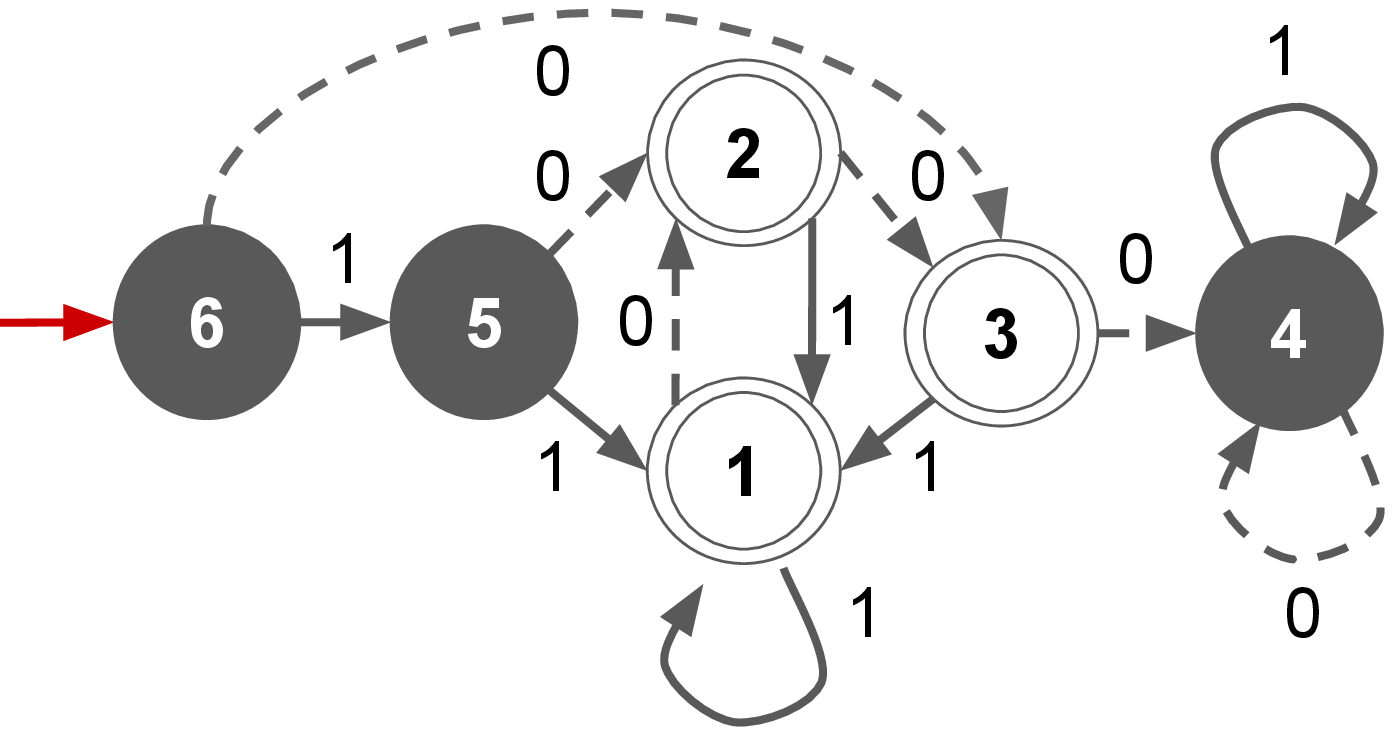}
  \caption{Incorrect DFA extracted on grammar 4.}
  \label{fig:Wrong_DFA_g4}
\end{subfigure} \hfill

\begin{subfigure}{.50\textwidth}
  \centering
  \includegraphics[width=1.0\linewidth]{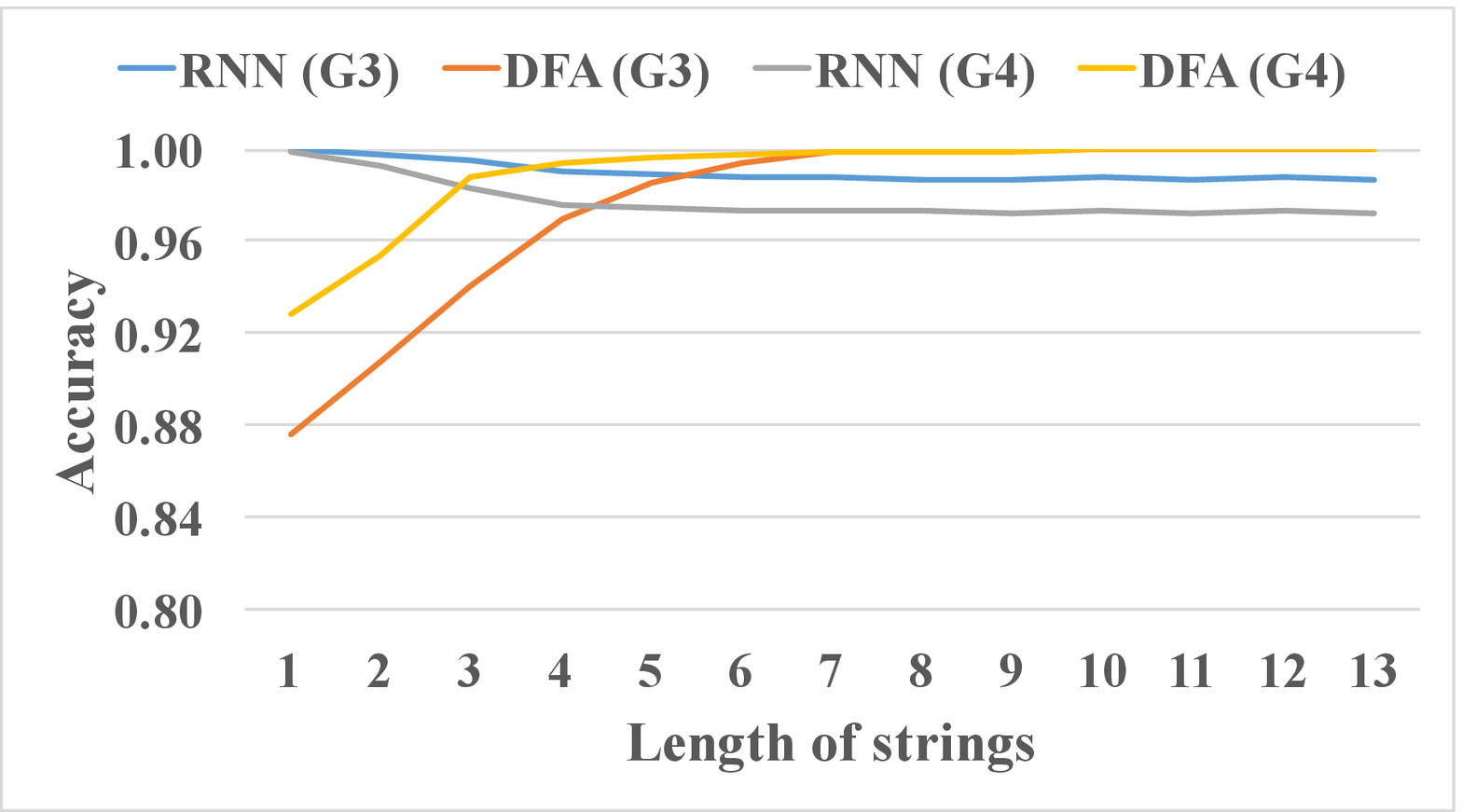}
  \caption{Comparing RNNs and DFA on testing \\
  sets comprised of longer strings.}
  \label{fig:rnn_vs_dfa}
\end{subfigure} \hfill%
\begin{subfigure}{.50\textwidth}
  \centering
  \includegraphics[width=1.0\linewidth]{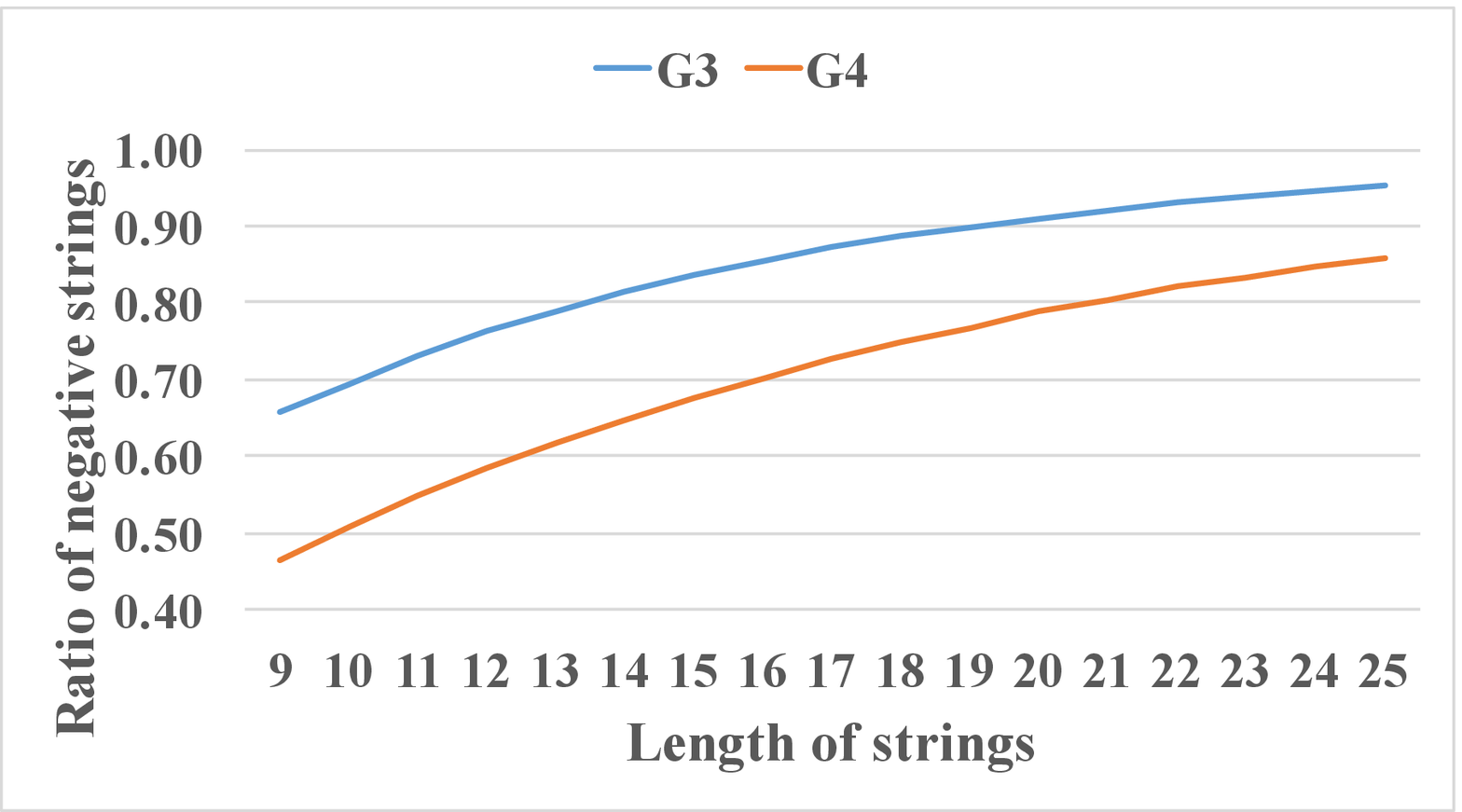}
  \caption{The ratio of negative strings in the \\ 
  complete set of strings of certain length.}
  \label{fig:neg_ratio}
\end{subfigure} \hfill
\caption{Evaluation of small-capacity RNNs and their associated incorrect DFA for Tomita grammars 3 and 4.}
\label{fig:practical_case}
\end{figure}

As discussed in Section~\ref{sec:model_complexity}, RNNs with larger capacity can learn a DFA better. In practice, it is usually not possible to know the appropriate capacity when constructing an RNN. As a result, it is possible that a smaller RNN can be well-trained on short sequences, but will generalize poorly when confronted with long sequences. Above experiments suggest one solution of extracting a DFA from a trained RNN model, given that DFA extraction is relatively stable and a DFA can maintain its accuracy of recognizing long strings. In reality, however, it is impractical to assume that the ground truth DFA can be obtained to evaluate the extracted ones, which are possible to be incorrect. In following experiments, we empirically compare some RNNs and their ``incorrectly'' extracted DFA. Here we demonstrate the results on grammar 3 and 4 due to space constraint. These grammars are selected as we observed in the experiments that the RNNs trained on these grammars are more sensitive to model capacity.

We first construct two RNNs with 9 hidden neurons and have them trained to reach 100\% accuracy on data set $D_{test}$. Their associated incorrect DFA extracted, as shown in Figure~\ref{fig:Wrong_DFA_g3} and~\ref{fig:Wrong_DFA_g4}, achieve 93\% and 98\% accuracy on $D_{test}$ respectively. We next evaluate these RNNs and their incorrect DFA using multiple testing sets with the number of samples fixed at 100,000 and string length varying from 20 to 200. The sampling of positive and negative strings is similar to what was described in Section.\ref{sec:data}. 

RNN test-set performance is shown in Figure.~\ref{fig:rnn_vs_dfa}. We observe that on these test sets composed of longer strings, RNNs make more classification errors. This may due to the fact that as the string length increases, the ratio of negative strings to positive ones also increases (shown in Figure.\ref{fig:neg_ratio}). This would mean that an RNN processes more negative strings, which can be interpreted as ``noisy'' samples, and as a result, would generate more false positive errors. On the other hand, for the DFA associated with these RNNs, fewer and fewer mistakes are made as the number of negative strings increases. This might be the result of the fact that these incorrect DFA generate their own regular language $L_{3'}$ and $L_{4'}$ respectively, which are quite similar to the target languages $L_{3}$ and $L_{4}$. As a result, many of the negative strings rejected by the extracted DFA are also rejected by the correct DFA. As more and more negative strings are sampled, this overlapping behavior gradually dominates the testing sets. These results demonstrate that, in certain cases, it is possible to extract a DFA which does not fully represent the target RNN and yet still outperforms the RNN when processing longer sequences. Given this result, one possible path to improving an RNN's ability to handle longer sequences might lie in the exploitation of this useful DFA behavior. 

\section{Conclusion}
We conducted a careful experimental study of the extraction of deterministic finite automata from second-order recurrent neural networks. We identified the factors that influence the reliability of the extraction process and were able to show that, despite these factors, the automata can still be stably extracted even when the neural model is trained only using short sequences. Our experiments also show that while model capacity does indeed strongly damage the neural network's ability to handle longer sequences, this hardly affects the extraction process. Furthermore, the automata extracted from low-capacity second order RNNs, in some cases, actually outperform the RNN trained model when processing sequences longer than what were seen during training. Our findings imply that one potential pathway to improving an RNN's ability to learn longer-term dependencies might be through the exploitation of the DFA's natural ability to handle infinitely long sequences and that it would be interesting to exploit transfer learning in this area. Future work will focus on comparing extracted DFAs and source RNN models on more complex/or real-world datasets consisting of long sequences such as currency exchange rates~\cite{giles2001noisy} and others.

\section{Acknowledgements}
We gratefully acknowledge partial support from the College of Information Sciences and Technology. 

\bibliographystyle{apacite}
\bibliography{ref}

\begin{thebibliography}{}

\bibitem [\protect \citeauthoryear {%
Borges%
, d'Avila Garcez%
\BCBL {}\ \BBA {} Lamb%
}{%
Borges%
\ \protect \BOthers {.}}{%
{\protect \APACyear {2011}}%
}]{%
BorgesGL11}
\APACinsertmetastar {%
BorgesGL11}%
\begin{APACrefauthors}%
Borges, R\BPBI V.%
, d'Avila Garcez, A\BPBI S.%
\BCBL {}\ \BBA {} Lamb, L\BPBI C.%
\end{APACrefauthors}%
\unskip\
\newblock
\APACrefYearMonthDay{2011}{}{}.
\newblock
{\BBOQ}\APACrefatitle {Learning and Representing Temporal Knowledge in
  Recurrent Networks} {Learning and representing temporal knowledge in
  recurrent networks}.{\BBCQ}
\newblock
\APACjournalVolNumPages{{IEEE} Trans. Neural Networks}{22}{12}{2409--2421}.
\PrintBackRefs{\CurrentBib}

\bibitem [\protect \citeauthoryear {%
Carroll%
\ \BBA {} Long%
}{%
Carroll%
\ \BBA {} Long%
}{%
{\protect \APACyear {1989}}%
}]{%
carroll1989theory}
\APACinsertmetastar {%
carroll1989theory}%
\begin{APACrefauthors}%
Carroll, J.%
\BCBT {}\ \BBA {} Long, D.%
\end{APACrefauthors}%
\unskip\
\newblock
\APACrefYear{1989}.
\newblock
\APACrefbtitle {Theory of finite automata with an introduction to formal
  languages} {Theory of finite automata with an introduction to formal
  languages}.
\PrintBackRefs{\CurrentBib}

\bibitem [\protect \citeauthoryear {%
Casey%
}{%
Casey%
}{%
{\protect \APACyear {1996}}%
}]{%
casey1996dynamics}
\APACinsertmetastar {%
casey1996dynamics}%
\begin{APACrefauthors}%
Casey, M.%
\end{APACrefauthors}%
\unskip\
\newblock
\APACrefYearMonthDay{1996}{}{}.
\newblock
{\BBOQ}\APACrefatitle {The dynamics of discrete-time computation, with
  application to recurrent neural networks and finite state machine extraction}
  {The dynamics of discrete-time computation, with application to recurrent
  neural networks and finite state machine extraction}.{\BBCQ}
\newblock
\APACjournalVolNumPages{Neural computation}{8}{6}{1135--1178}.
\PrintBackRefs{\CurrentBib}

\bibitem [\protect \citeauthoryear {%
Cho%
, Van~Merri{\"e}nboer%
, Bahdanau%
\BCBL {}\ \BBA {} Bengio%
}{%
Cho%
\ \protect \BOthers {.}}{%
{\protect \APACyear {2014}}%
}]{%
cho2014properties}
\APACinsertmetastar {%
cho2014properties}%
\begin{APACrefauthors}%
Cho, K.%
, Van~Merri{\"e}nboer, B.%
, Bahdanau, D.%
\BCBL {}\ \BBA {} Bengio, Y.%
\end{APACrefauthors}%
\unskip\
\newblock
\APACrefYearMonthDay{2014}{}{}.
\newblock
{\BBOQ}\APACrefatitle {On the properties of neural machine translation:
  Encoder-decoder approaches} {On the properties of neural machine translation:
  Encoder-decoder approaches}.{\BBCQ}
\newblock
\APACjournalVolNumPages{arXiv preprint arXiv:1409.1259}{}{}{}.
\PrintBackRefs{\CurrentBib}

\bibitem [\protect \citeauthoryear {%
Chomsky%
}{%
Chomsky%
}{%
{\protect \APACyear {1956}}%
}]{%
chomsky1956three}
\APACinsertmetastar {%
chomsky1956three}%
\begin{APACrefauthors}%
Chomsky, N.%
\end{APACrefauthors}%
\unskip\
\newblock
\APACrefYearMonthDay{1956}{}{}.
\newblock
{\BBOQ}\APACrefatitle {Three models for the description of language} {Three
  models for the description of language}.{\BBCQ}
\newblock
\APACjournalVolNumPages{IRE Transactions on information
  theory}{2}{3}{113--124}.
\PrintBackRefs{\CurrentBib}

\bibitem [\protect \citeauthoryear {%
Dhingra%
, Yang%
, Cohen%
\BCBL {}\ \BBA {} Salakhutdinov%
}{%
Dhingra%
\ \protect \BOthers {.}}{%
{\protect \APACyear {2017}}%
}]{%
dhingra2017linguistic}
\APACinsertmetastar {%
dhingra2017linguistic}%
\begin{APACrefauthors}%
Dhingra, B.%
, Yang, Z.%
, Cohen, W\BPBI W.%
\BCBL {}\ \BBA {} Salakhutdinov, R.%
\end{APACrefauthors}%
\unskip\
\newblock
\APACrefYearMonthDay{2017}{}{}.
\newblock
{\BBOQ}\APACrefatitle {Linguistic Knowledge as Memory for Recurrent Neural
  Networks} {Linguistic knowledge as memory for recurrent neural
  networks}.{\BBCQ}
\newblock
\APACjournalVolNumPages{arXiv preprint arXiv:1703.02620}{}{}{}.
\PrintBackRefs{\CurrentBib}

\bibitem [\protect \citeauthoryear {%
Du%
, Ma%
, Wu%
, Kar%
\BCBL {}\ \BBA {} Moura%
}{%
Du%
\ \protect \BOthers {.}}{%
{\protect \APACyear {2016}}%
}]{%
du2016convergence}
\APACinsertmetastar {%
du2016convergence}%
\begin{APACrefauthors}%
Du, J.%
, Ma, S.%
, Wu, Y\BHBI C.%
, Kar, S.%
\BCBL {}\ \BBA {} Moura, J\BPBI M.%
\end{APACrefauthors}%
\unskip\
\newblock
\APACrefYearMonthDay{2016}{}{}.
\newblock
{\BBOQ}\APACrefatitle {Convergence analysis of distributed inference with
  vector-valued Gaussian belief propagation} {Convergence analysis of
  distributed inference with vector-valued gaussian belief propagation}.{\BBCQ}
\newblock
\APACjournalVolNumPages{arXiv preprint arXiv:1611.02010}{}{}{}.
\PrintBackRefs{\CurrentBib}

\bibitem [\protect \citeauthoryear {%
Du%
, Zhang%
, Wu%
, Moura%
\BCBL {}\ \BBA {} Kar%
}{%
Du%
\ \protect \BOthers {.}}{%
{\protect \APACyear {2017}}%
}]{%
du2017topology}
\APACinsertmetastar {%
du2017topology}%
\begin{APACrefauthors}%
Du, J.%
, Zhang, S.%
, Wu, G.%
, Moura, J\BPBI M.%
\BCBL {}\ \BBA {} Kar, S.%
\end{APACrefauthors}%
\unskip\
\newblock
\APACrefYearMonthDay{2017}{}{}.
\newblock
{\BBOQ}\APACrefatitle {Topology adaptive graph convolutional networks}
  {Topology adaptive graph convolutional networks}.{\BBCQ}
\newblock
\APACjournalVolNumPages{arXiv preprint arXiv:1710.10370}{}{}{}.
\PrintBackRefs{\CurrentBib}

\bibitem [\protect \citeauthoryear {%
Elman%
}{%
Elman%
}{%
{\protect \APACyear {1990}}%
}]{%
elman1990finding}
\APACinsertmetastar {%
elman1990finding}%
\begin{APACrefauthors}%
Elman, J\BPBI L.%
\end{APACrefauthors}%
\unskip\
\newblock
\APACrefYearMonthDay{1990}{}{}.
\newblock
{\BBOQ}\APACrefatitle {Finding structure in time} {Finding structure in
  time}.{\BBCQ}
\newblock
\APACjournalVolNumPages{Cognitive science}{14}{2}{179--211}.
\PrintBackRefs{\CurrentBib}

\bibitem [\protect \citeauthoryear {%
Frasconi%
, Gori%
, Maggini%
\BCBL {}\ \BBA {} Soda%
}{%
Frasconi%
\ \protect \BOthers {.}}{%
{\protect \APACyear {1996}}%
}]{%
frasconi1996representation}
\APACinsertmetastar {%
frasconi1996representation}%
\begin{APACrefauthors}%
Frasconi, P.%
, Gori, M.%
, Maggini, M.%
\BCBL {}\ \BBA {} Soda, G.%
\end{APACrefauthors}%
\unskip\
\newblock
\APACrefYearMonthDay{1996}{}{}.
\newblock
{\BBOQ}\APACrefatitle {Representation of finite state automata in recurrent
  radial basis function networks} {Representation of finite state automata in
  recurrent radial basis function networks}.{\BBCQ}
\newblock
\APACjournalVolNumPages{Machine Learning}{23}{1}{5--32}.
\PrintBackRefs{\CurrentBib}

\bibitem [\protect \citeauthoryear {%
Giles%
\ \protect \BOthers {.}}{%
Giles%
\ \protect \BOthers {.}}{%
{\protect \APACyear {1991}}%
}]{%
giles1991second}
\APACinsertmetastar {%
giles1991second}%
\begin{APACrefauthors}%
Giles, C\BPBI L.%
, Chen, D.%
, Miller, C.%
, Chen, H.%
, Sun, G.%
\BCBL {}\ \BBA {} Lee, Y.%
\end{APACrefauthors}%
\unskip\
\newblock
\APACrefYearMonthDay{1991}{}{}.
\newblock
{\BBOQ}\APACrefatitle {Second-order recurrent neural networks for grammatical
  inference} {Second-order recurrent neural networks for grammatical
  inference}.{\BBCQ}
\newblock
\BIn{} \APACrefbtitle {Neural Networks, 1991., IJCNN-91-Seattle International
  Joint Conference on} {Neural networks, 1991., ijcnn-91-seattle international
  joint conference on}\ (\BVOL~2, \BPGS\ 273--281).
\PrintBackRefs{\CurrentBib}

\bibitem [\protect \citeauthoryear {%
Giles%
, Lawrence%
\BCBL {}\ \BBA {} Tsoi%
}{%
Giles%
\ \protect \BOthers {.}}{%
{\protect \APACyear {2001}}%
}]{%
giles2001noisy}
\APACinsertmetastar {%
giles2001noisy}%
\begin{APACrefauthors}%
Giles, C\BPBI L.%
, Lawrence, S.%
\BCBL {}\ \BBA {} Tsoi, A\BPBI C.%
\end{APACrefauthors}%
\unskip\
\newblock
\APACrefYearMonthDay{2001}{}{}.
\newblock
{\BBOQ}\APACrefatitle {Noisy time series prediction using recurrent neural
  networks and grammatical inference} {Noisy time series prediction using
  recurrent neural networks and grammatical inference}.{\BBCQ}
\newblock
\APACjournalVolNumPages{Machine learning}{44}{1-2}{161--183}.
\PrintBackRefs{\CurrentBib}

\bibitem [\protect \citeauthoryear {%
Giles%
\ \protect \BOthers {.}}{%
Giles%
\ \protect \BOthers {.}}{%
{\protect \APACyear {1992}}%
}]{%
giles1992learning}
\APACinsertmetastar {%
giles1992learning}%
\begin{APACrefauthors}%
Giles, C\BPBI L.%
, Miller, C\BPBI B.%
, Chen, D.%
, Chen, H\BHBI H.%
, Sun, G\BHBI Z.%
\BCBL {}\ \BBA {} Lee, Y\BHBI C.%
\end{APACrefauthors}%
\unskip\
\newblock
\APACrefYearMonthDay{1992}{}{}.
\newblock
{\BBOQ}\APACrefatitle {Learning and extracting finite state automata with
  second-order recurrent neural networks} {Learning and extracting finite state
  automata with second-order recurrent neural networks}.{\BBCQ}
\newblock
\APACjournalVolNumPages{Neural Computation}{4}{3}{393--405}.
\PrintBackRefs{\CurrentBib}

\bibitem [\protect \citeauthoryear {%
Giles%
, Sun%
, Chen%
, Lee%
\BCBL {}\ \BBA {} Chen%
}{%
Giles%
\ \protect \BOthers {.}}{%
{\protect \APACyear {1990}}%
}]{%
giles1990higher}
\APACinsertmetastar {%
giles1990higher}%
\begin{APACrefauthors}%
Giles, C\BPBI L.%
, Sun, G\BHBI Z.%
, Chen, H\BHBI H.%
, Lee, Y\BHBI C.%
\BCBL {}\ \BBA {} Chen, D.%
\end{APACrefauthors}%
\unskip\
\newblock
\APACrefYearMonthDay{1990}{}{}.
\newblock
{\BBOQ}\APACrefatitle {Higher order recurrent networks and grammatical
  inference} {Higher order recurrent networks and grammatical
  inference}.{\BBCQ}
\newblock
\BIn{} \APACrefbtitle {Advances in neural information processing systems}
  {Advances in neural information processing systems}\ (\BPGS\ 380--387).
\PrintBackRefs{\CurrentBib}

\bibitem [\protect \citeauthoryear {%
Gori%
, Maggini%
, Martinelli%
\BCBL {}\ \BBA {} Soda%
}{%
Gori%
\ \protect \BOthers {.}}{%
{\protect \APACyear {1998}}%
}]{%
gori1998inductive}
\APACinsertmetastar {%
gori1998inductive}%
\begin{APACrefauthors}%
Gori, M.%
, Maggini, M.%
, Martinelli, E.%
\BCBL {}\ \BBA {} Soda, G.%
\end{APACrefauthors}%
\unskip\
\newblock
\APACrefYearMonthDay{1998}{}{}.
\newblock
{\BBOQ}\APACrefatitle {Inductive inference from noisy examples using the hybrid
  finite state filter} {Inductive inference from noisy examples using the
  hybrid finite state filter}.{\BBCQ}
\newblock
\APACjournalVolNumPages{IEEE Transactions on Neural Networks}{9}{3}{571--575}.
\PrintBackRefs{\CurrentBib}

\bibitem [\protect \citeauthoryear {%
Hochreiter%
\ \BBA {} Schmidhuber%
}{%
Hochreiter%
\ \BBA {} Schmidhuber%
}{%
{\protect \APACyear {1997}}%
}]{%
hochreiter1997long}
\APACinsertmetastar {%
hochreiter1997long}%
\begin{APACrefauthors}%
Hochreiter, S.%
\BCBT {}\ \BBA {} Schmidhuber, J.%
\end{APACrefauthors}%
\unskip\
\newblock
\APACrefYearMonthDay{1997}{}{}.
\newblock
{\BBOQ}\APACrefatitle {Long short-term memory} {Long short-term memory}.{\BBCQ}
\newblock
\APACjournalVolNumPages{Neural computation}{9}{8}{1735--1780}.
\PrintBackRefs{\CurrentBib}

\bibitem [\protect \citeauthoryear {%
Hopcroft%
, Motwani%
\BCBL {}\ \BBA {} Ullman%
}{%
Hopcroft%
\ \protect \BOthers {.}}{%
{\protect \APACyear {2013}}%
}]{%
hopcroft2006automata}
\APACinsertmetastar {%
hopcroft2006automata}%
\begin{APACrefauthors}%
Hopcroft, J\BPBI E.%
, Motwani, R.%
\BCBL {}\ \BBA {} Ullman, J\BPBI D.%
\end{APACrefauthors}%
\unskip\
\newblock
\APACrefYear{2013}.
\newblock
\APACrefbtitle {Introduction to Automata theory, Languages, and Computation}
  {Introduction to automata theory, languages, and computation}.
\PrintBackRefs{\CurrentBib}

\bibitem [\protect \citeauthoryear {%
Jacobsson%
}{%
Jacobsson%
}{%
{\protect \APACyear {2005}}%
}]{%
jacobsson2005rule}
\APACinsertmetastar {%
jacobsson2005rule}%
\begin{APACrefauthors}%
Jacobsson, H.%
\end{APACrefauthors}%
\unskip\
\newblock
\APACrefYearMonthDay{2005}{}{}.
\newblock
{\BBOQ}\APACrefatitle {Rule extraction from recurrent neural networks: a
  taxonomy and review} {Rule extraction from recurrent neural networks: a
  taxonomy and review}.{\BBCQ}
\newblock
\APACjournalVolNumPages{Neural Computation}{17}{6}{1223--1263}.
\PrintBackRefs{\CurrentBib}

\bibitem [\protect \citeauthoryear {%
Kolen%
}{%
Kolen%
}{%
{\protect \APACyear {1994}}%
}]{%
kolen1994fool}
\APACinsertmetastar {%
kolen1994fool}%
\begin{APACrefauthors}%
Kolen, J\BPBI F.%
\end{APACrefauthors}%
\unskip\
\newblock
\APACrefYearMonthDay{1994}{}{}.
\newblock
{\BBOQ}\APACrefatitle {Fool's gold: Extracting finite state machines from
  recurrent network dynamics} {Fool's gold: Extracting finite state machines
  from recurrent network dynamics}.{\BBCQ}
\newblock
\BIn{} \APACrefbtitle {Advances in neural information processing systems}
  {Advances in neural information processing systems}\ (\BPGS\ 501--508).
\PrintBackRefs{\CurrentBib}

\bibitem [\protect \citeauthoryear {%
Li%
\ \BBA {} Pr{\'{\i}}ncipe%
}{%
Li%
\ \BBA {} Pr{\'{\i}}ncipe%
}{%
{\protect \APACyear {2016}}%
}]{%
principe2016}
\APACinsertmetastar {%
principe2016}%
\begin{APACrefauthors}%
Li, K.%
\BCBT {}\ \BBA {} Pr{\'{\i}}ncipe, J\BPBI C.%
\end{APACrefauthors}%
\unskip\
\newblock
\APACrefYearMonthDay{2016}{}{}.
\newblock
{\BBOQ}\APACrefatitle {The Kernel Adaptive Autoregressive-Moving-Average
  Algorithm} {The kernel adaptive autoregressive-moving-average
  algorithm}.{\BBCQ}
\newblock
\APACjournalVolNumPages{{IEEE} Trans. Neural Netw. Learning
  Syst.}{27}{2}{334--346}.
\PrintBackRefs{\CurrentBib}

\bibitem [\protect \citeauthoryear {%
Lin%
, Horne%
, Ti{\~{n}}o%
\BCBL {}\ \BBA {} Giles%
}{%
Lin%
\ \protect \BOthers {.}}{%
{\protect \APACyear {1996}}%
}]{%
Lin96NARX}
\APACinsertmetastar {%
Lin96NARX}%
\begin{APACrefauthors}%
Lin, T.%
, Horne, B\BPBI G.%
, Ti{\~{n}}o, P.%
\BCBL {}\ \BBA {} Giles, C\BPBI L.%
\end{APACrefauthors}%
\unskip\
\newblock
\APACrefYearMonthDay{1996}{}{}.
\newblock
{\BBOQ}\APACrefatitle {Learning long-term dependencies in {NARX} recurrent
  neural networks} {Learning long-term dependencies in {NARX} recurrent neural
  networks}.{\BBCQ}
\newblock
\APACjournalVolNumPages{{IEEE} Trans. Neural Networks}{7}{6}{1329--1338}.
\PrintBackRefs{\CurrentBib}

\bibitem [\protect \citeauthoryear {%
Minsky%
}{%
Minsky%
}{%
{\protect \APACyear {1967}}%
}]{%
minsky1967computation}
\APACinsertmetastar {%
minsky1967computation}%
\begin{APACrefauthors}%
Minsky, M\BPBI L.%
\end{APACrefauthors}%
\unskip\
\newblock
\APACrefYear{1967}.
\newblock
\APACrefbtitle {Computation: finite and infinite machines} {Computation: finite
  and infinite machines}.
\newblock
\APACaddressPublisher{}{Prentice-Hall, Inc.}
\PrintBackRefs{\CurrentBib}

\bibitem [\protect \citeauthoryear {%
Murdoch%
\ \BBA {} Szlam%
}{%
Murdoch%
\ \BBA {} Szlam%
}{%
{\protect \APACyear {2017}}%
}]{%
murdoch2017automatic}
\APACinsertmetastar {%
murdoch2017automatic}%
\begin{APACrefauthors}%
Murdoch, W\BPBI J.%
\BCBT {}\ \BBA {} Szlam, A.%
\end{APACrefauthors}%
\unskip\
\newblock
\APACrefYearMonthDay{2017}{}{}.
\newblock
{\BBOQ}\APACrefatitle {Automatic Rule Extraction from Long Short Term Memory
  Networks} {Automatic rule extraction from long short term memory
  networks}.{\BBCQ}
\newblock
\APACjournalVolNumPages{arXiv preprint arXiv:1702.02540}{}{}{}.
\PrintBackRefs{\CurrentBib}

\bibitem [\protect \citeauthoryear {%
Omlin%
\ \BBA {} Giles%
}{%
Omlin%
\ \BBA {} Giles%
}{%
{\protect \APACyear {1996}}%
{\protect \APACexlab {{\protect \BCnt {1}}}}}]{%
omlin1996jacm}
\APACinsertmetastar {%
omlin1996jacm}%
\begin{APACrefauthors}%
Omlin, C\BPBI W.%
\BCBT {}\ \BBA {} Giles, C\BPBI L.%
\end{APACrefauthors}%
\unskip\
\newblock
\APACrefYearMonthDay{1996{\protect \BCnt {1}}}{}{}.
\newblock
{\BBOQ}\APACrefatitle {Constructing deterministic finite-state automata in
  recurrent neural networks} {Constructing deterministic finite-state automata
  in recurrent neural networks}.{\BBCQ}
\newblock
\APACjournalVolNumPages{Journal of the ACM}{43}{6}{937--972}.
\PrintBackRefs{\CurrentBib}

\bibitem [\protect \citeauthoryear {%
Omlin%
\ \BBA {} Giles%
}{%
Omlin%
\ \BBA {} Giles%
}{%
{\protect \APACyear {1996}}%
{\protect \APACexlab {{\protect \BCnt {2}}}}}]{%
omlin1996stable}
\APACinsertmetastar {%
omlin1996stable}%
\begin{APACrefauthors}%
Omlin, C\BPBI W.%
\BCBT {}\ \BBA {} Giles, C\BPBI L.%
\end{APACrefauthors}%
\unskip\
\newblock
\APACrefYearMonthDay{1996{\protect \BCnt {2}}}{}{}.
\newblock
{\BBOQ}\APACrefatitle {Constructing deterministic finite-state automata in
  recurrent neural networks} {Constructing deterministic finite-state automata
  in recurrent neural networks}.{\BBCQ}
\newblock
\APACjournalVolNumPages{Neural Computation}{8}{4}{675--696}.
\PrintBackRefs{\CurrentBib}

\bibitem [\protect \citeauthoryear {%
Omlin%
\ \BBA {} Giles%
}{%
Omlin%
\ \BBA {} Giles%
}{%
{\protect \APACyear {1996}}%
{\protect \APACexlab {{\protect \BCnt {3}}}}}]{%
omlin1996extraction}
\APACinsertmetastar {%
omlin1996extraction}%
\begin{APACrefauthors}%
Omlin, C\BPBI W.%
\BCBT {}\ \BBA {} Giles, C\BPBI L.%
\end{APACrefauthors}%
\unskip\
\newblock
\APACrefYearMonthDay{1996{\protect \BCnt {3}}}{}{}.
\newblock
{\BBOQ}\APACrefatitle {Extraction of rules from discrete-time recurrent neural
  networks} {Extraction of rules from discrete-time recurrent neural
  networks}.{\BBCQ}
\newblock
\APACjournalVolNumPages{Neural networks}{9}{1}{41--52}.
\PrintBackRefs{\CurrentBib}

\bibitem [\protect \citeauthoryear {%
Omlin%
\ \BBA {} Giles%
}{%
Omlin%
\ \BBA {} Giles%
}{%
{\protect \APACyear {2000}}%
}]{%
omlin2000symbolic}
\APACinsertmetastar {%
omlin2000symbolic}%
\begin{APACrefauthors}%
Omlin, C\BPBI W.%
\BCBT {}\ \BBA {} Giles, C\BPBI L.%
\end{APACrefauthors}%
\unskip\
\newblock
\APACrefYearMonthDay{2000}{}{}.
\newblock
{\BBOQ}\APACrefatitle {Symbolic knowledge representation in recurrent neural
  networks: Insights from theoretical models of computation} {Symbolic
  knowledge representation in recurrent neural networks: Insights from
  theoretical models of computation}.{\BBCQ}
\newblock
\APACjournalVolNumPages{Knowledge based neurocomputing}{}{}{63--115}.
\PrintBackRefs{\CurrentBib}

\bibitem [\protect \citeauthoryear {%
Pascanu%
, Mikolov%
\BCBL {}\ \BBA {} Bengio%
}{%
Pascanu%
\ \protect \BOthers {.}}{%
{\protect \APACyear {2013}}%
}]{%
pascanu2013difficulty}
\APACinsertmetastar {%
pascanu2013difficulty}%
\begin{APACrefauthors}%
Pascanu, R.%
, Mikolov, T.%
\BCBL {}\ \BBA {} Bengio, Y.%
\end{APACrefauthors}%
\unskip\
\newblock
\APACrefYearMonthDay{2013}{}{}.
\newblock
{\BBOQ}\APACrefatitle {On the difficulty of training recurrent neural networks}
  {On the difficulty of training recurrent neural networks}.{\BBCQ}
\newblock
\BIn{} \APACrefbtitle {International Conference on Machine Learning}
  {International conference on machine learning}\ (\BPGS\ 1310--1318).
\PrintBackRefs{\CurrentBib}

\bibitem [\protect \citeauthoryear {%
Pollack%
}{%
Pollack%
}{%
{\protect \APACyear {1991}}%
}]{%
pollack1991induction}
\APACinsertmetastar {%
pollack1991induction}%
\begin{APACrefauthors}%
Pollack, J\BPBI B.%
\end{APACrefauthors}%
\unskip\
\newblock
\APACrefYearMonthDay{1991}{}{}.
\newblock
{\BBOQ}\APACrefatitle {The induction of dynamical recognizers} {The induction
  of dynamical recognizers}.{\BBCQ}
\newblock
\APACjournalVolNumPages{Machine learning}{7}{2}{227--252}.
\PrintBackRefs{\CurrentBib}

\bibitem [\protect \citeauthoryear {%
Sanfeliu%
\ \BBA {} Alquezar%
}{%
Sanfeliu%
\ \BBA {} Alquezar%
}{%
{\protect \APACyear {1994}}%
}]{%
sanfeliu1994active}
\APACinsertmetastar {%
sanfeliu1994active}%
\begin{APACrefauthors}%
Sanfeliu, A.%
\BCBT {}\ \BBA {} Alquezar, R.%
\end{APACrefauthors}%
\unskip\
\newblock
\APACrefYearMonthDay{1994}{}{}.
\newblock
{\BBOQ}\APACrefatitle {Active grammatical inference: a new learning
  methodology} {Active grammatical inference: a new learning
  methodology}.{\BBCQ}
\newblock
\BIn{} \APACrefbtitle {in Shape, Structure and Pattern Recogniton, D. Dori and
  A. Bruckstein (eds.), World Scientific Pub.} {in shape, structure and pattern
  recogniton, d. dori and a. bruckstein (eds.), world scientific pub.}
\PrintBackRefs{\CurrentBib}

\bibitem [\protect \citeauthoryear {%
Schellhammer%
, Diederich%
, Towsey%
\BCBL {}\ \BBA {} Brugman%
}{%
Schellhammer%
\ \protect \BOthers {.}}{%
{\protect \APACyear {1998}}%
}]{%
schellhammer1998knowledge}
\APACinsertmetastar {%
schellhammer1998knowledge}%
\begin{APACrefauthors}%
Schellhammer, I.%
, Diederich, J.%
, Towsey, M.%
\BCBL {}\ \BBA {} Brugman, C.%
\end{APACrefauthors}%
\unskip\
\newblock
\APACrefYearMonthDay{1998}{}{}.
\newblock
{\BBOQ}\APACrefatitle {Knowledge extraction and recurrent neural networks: An
  analysis of an Elman network trained on a natural language learning task}
  {Knowledge extraction and recurrent neural networks: An analysis of an elman
  network trained on a natural language learning task}.{\BBCQ}
\newblock
\BIn{} \APACrefbtitle {Proceedings of the Joint Conferences on New Methods in
  Language Processing and Computational Natural Language Learning} {Proceedings
  of the joint conferences on new methods in language processing and
  computational natural language learning}\ (\BPGS\ 73--78).
\PrintBackRefs{\CurrentBib}

\bibitem [\protect \citeauthoryear {%
Sukhbaatar%
, Weston%
, Fergus%
\BCBL {}\ \protect \BOthers {.}}{%
Sukhbaatar%
\ \protect \BOthers {.}}{%
{\protect \APACyear {2015}}%
}]{%
sukhbaatar2015end}
\APACinsertmetastar {%
sukhbaatar2015end}%
\begin{APACrefauthors}%
Sukhbaatar, S.%
, Weston, J.%
, Fergus, R.%
\BCBL {}\ \BOthersPeriod {.}\end{APACrefauthors}%
\unskip\
\newblock
\APACrefYearMonthDay{2015}{}{}.
\newblock
{\BBOQ}\APACrefatitle {End-to-end memory networks} {End-to-end memory
  networks}.{\BBCQ}
\newblock
\BIn{} \APACrefbtitle {Advances in neural information processing systems}
  {Advances in neural information processing systems}\ (\BPGS\ 2440--2448).
\PrintBackRefs{\CurrentBib}

\bibitem [\protect \citeauthoryear {%
Tieleman%
\ \BBA {} Hinton%
}{%
Tieleman%
\ \BBA {} Hinton%
}{%
{\protect \APACyear {2012}}%
}]{%
tieleman2012lecture}
\APACinsertmetastar {%
tieleman2012lecture}%
\begin{APACrefauthors}%
Tieleman, T.%
\BCBT {}\ \BBA {} Hinton, G.%
\end{APACrefauthors}%
\unskip\
\newblock
\APACrefYearMonthDay{2012}{}{}.
\newblock
{\BBOQ}\APACrefatitle {Lecture 6.5-rmsprop: Divide the gradient by a running
  average of its recent magnitude} {Lecture 6.5-rmsprop: Divide the gradient by
  a running average of its recent magnitude}.{\BBCQ}
\newblock
\APACjournalVolNumPages{COURSERA: Neural networks for machine
  learning}{4}{2}{26--31}.
\PrintBackRefs{\CurrentBib}

\bibitem [\protect \citeauthoryear {%
Ti{\v{n}}o%
\ \BBA {} {\v{S}}ajda%
}{%
Ti{\v{n}}o%
\ \BBA {} {\v{S}}ajda%
}{%
{\protect \APACyear {1995}}%
}]{%
tivno1995learning}
\APACinsertmetastar {%
tivno1995learning}%
\begin{APACrefauthors}%
Ti{\v{n}}o, P.%
\BCBT {}\ \BBA {} {\v{S}}ajda, J.%
\end{APACrefauthors}%
\unskip\
\newblock
\APACrefYearMonthDay{1995}{}{}.
\newblock
{\BBOQ}\APACrefatitle {Learning and extracting initial mealy automata with a
  modular neural network model} {Learning and extracting initial mealy automata
  with a modular neural network model}.{\BBCQ}
\newblock
\APACjournalVolNumPages{Neural Computation}{7}{4}{822--844}.
\PrintBackRefs{\CurrentBib}

\bibitem [\protect \citeauthoryear {%
Tomita%
}{%
Tomita%
}{%
{\protect \APACyear {1982}}%
{\protect \APACexlab {{\protect \BCnt {1}}}}}]{%
tomita1982}
\APACinsertmetastar {%
tomita1982}%
\begin{APACrefauthors}%
Tomita, M.%
\end{APACrefauthors}%
\unskip\
\newblock
\APACrefYearMonthDay{1982{\protect \BCnt {1}}}{}{}.
\newblock
{\BBOQ}\APACrefatitle {Dynamic construction of finite automata from example
  using hill-climbing} {Dynamic construction of finite automata from example
  using hill-climbing}.{\BBCQ}
\newblock
\APACjournalVolNumPages{Proceedings of the Fourth Annual Cognitive Science
  Conference}{}{}{105--108}.
\PrintBackRefs{\CurrentBib}

\bibitem [\protect \citeauthoryear {%
Tomita%
}{%
Tomita%
}{%
{\protect \APACyear {1982}}%
{\protect \APACexlab {{\protect \BCnt {2}}}}}]{%
tomita1992dynamic}
\APACinsertmetastar {%
tomita1992dynamic}%
\begin{APACrefauthors}%
Tomita, M.%
\end{APACrefauthors}%
\unskip\
\newblock
\APACrefYearMonthDay{1982{\protect \BCnt {2}}}{}{}.
\newblock
{\BBOQ}\APACrefatitle {Dynamic Construction of Finite-State Automata from
  Examples using Hill-Climbing} {Dynamic construction of finite-state automata
  from examples using hill-climbing}.{\BBCQ}
\newblock
\BIn{} (\BPGS\ 105--108).
\PrintBackRefs{\CurrentBib}

\bibitem [\protect \citeauthoryear {%
Wang%
\ \protect \BOthers {.}}{%
Wang%
\ \protect \BOthers {.}}{%
{\protect \APACyear {2018}}%
}]{%
wang2018model}
\APACinsertmetastar {%
wang2018model}%
\begin{APACrefauthors}%
Wang, Q.%
, Zhang, K.%
, II, A\BPBI G\BPBI O.%
, Xing, X.%
, Liu, X.%
\BCBL {}\ \BBA {} Giles, C\BPBI L.%
\end{APACrefauthors}%
\unskip\
\newblock
\APACrefYearMonthDay{2018}{}{}.
\newblock
{\BBOQ}\APACrefatitle {A Comparison of Rule Extraction for Different Recurrent
  Neural Network Models and Grammatical Complexity} {A comparison of rule
  extraction for different recurrent neural network models and grammatical
  complexity}.{\BBCQ}
\newblock
\APACjournalVolNumPages{arXiv preprint arXiv:1801.05420}{}{}{}.
\PrintBackRefs{\CurrentBib}

\bibitem [\protect \citeauthoryear {%
Watrous%
\ \BBA {} Kuhn%
}{%
Watrous%
\ \BBA {} Kuhn%
}{%
{\protect \APACyear {1992}}%
}]{%
watrous1992induction}
\APACinsertmetastar {%
watrous1992induction}%
\begin{APACrefauthors}%
Watrous, R\BPBI L.%
\BCBT {}\ \BBA {} Kuhn, G\BPBI M.%
\end{APACrefauthors}%
\unskip\
\newblock
\APACrefYearMonthDay{1992}{}{}.
\newblock
{\BBOQ}\APACrefatitle {Induction of finite-state automata using second-order
  recurrent networks} {Induction of finite-state automata using second-order
  recurrent networks}.{\BBCQ}
\newblock
\BIn{} \APACrefbtitle {Advances in neural information processing systems}
  {Advances in neural information processing systems}\ (\BPGS\ 309--317).
\PrintBackRefs{\CurrentBib}

\bibitem [\protect \citeauthoryear {%
Weston%
, Chopra%
\BCBL {}\ \BBA {} Bordes%
}{%
Weston%
\ \protect \BOthers {.}}{%
{\protect \APACyear {2014}}%
}]{%
weston2014memory}
\APACinsertmetastar {%
weston2014memory}%
\begin{APACrefauthors}%
Weston, J.%
, Chopra, S.%
\BCBL {}\ \BBA {} Bordes, A.%
\end{APACrefauthors}%
\unskip\
\newblock
\APACrefYearMonthDay{2014}{}{}.
\newblock
{\BBOQ}\APACrefatitle {Memory networks} {Memory networks}.{\BBCQ}
\newblock
\APACjournalVolNumPages{arXiv preprint arXiv:1410.3916}{}{}{}.
\PrintBackRefs{\CurrentBib}

\bibitem [\protect \citeauthoryear {%
Zeng%
, Goodman%
\BCBL {}\ \BBA {} Smyth%
}{%
Zeng%
\ \protect \BOthers {.}}{%
{\protect \APACyear {1993}}%
}]{%
zeng1993learning}
\APACinsertmetastar {%
zeng1993learning}%
\begin{APACrefauthors}%
Zeng, Z.%
, Goodman, R\BPBI M.%
\BCBL {}\ \BBA {} Smyth, P.%
\end{APACrefauthors}%
\unskip\
\newblock
\APACrefYearMonthDay{1993}{}{}.
\newblock
{\BBOQ}\APACrefatitle {Learning finite state machines with self-clustering
  recurrent networks} {Learning finite state machines with self-clustering
  recurrent networks}.{\BBCQ}
\newblock
\APACjournalVolNumPages{Neural Computation}{5}{6}{976--990}.
\PrintBackRefs{\CurrentBib}

\end{thebibliography}

\newpage
\appendix
\section*{Experimental Results for All Tomita Grammars}

\begin{table}[h]
\small
\centering
\caption{Influence of training time on DFA extraction and RNN performance for all Tomita grammars.}
\small
\begin{tabular}{ccccccccc}
\hline \hline
Grammar             & \multicolumn{8}{c}{Classification errors reached under different training epochs.} \\ \hline \hline
\multirow{4}{*}{G1} & Epoch     & 30        & 60        & 90       & 120       & 150     & 180     & 210     \\ \cline{2-9} 
                    & RNN(Train)       & 0.0      & 0.0        & 0.98       & 1.0     & 1.0     & 1.0      & 1.0      \\ \cline{2-9} 
                    & RNN(Test)       & 0.0      & 0.0        & 4.2e-3       & 4.3e-3     & 4.3e-3     & 4.4e-3      & 4.4e-3      \\ \cline{2-9} 
                    & DFA(Test)       & 4.0e-3     & 4.0e-3         & 1.0        & 1.0        & 1.0      & 1.0      & 1.0      \\ \hline \hline
\multirow{4}{*}{G2} & Epoch     & 100        & 200        & 300       & 400       & 500     & 600     & 700     \\ \cline{2-9} 
                    & RNN(Train)       & 0.0      & 0.54        & 0.91       & 0.85     & 0.85     & 0.92      & 0.97      \\ \cline{2-9}
                    & RNN(Test)       & 0.0      & 0.0        & 2.0e-2       & 2.0e-2     & 2.0e-2     & 2.0e-2      & 2.0e-2      \\ \cline{2-9}
                    & DFA       & 2.0e-3     & 2.0e-3         & 1.0        & 1.0        & 1.0      & 1.0      & 1.0      \\ \hline \hline
\multirow{4}{*}{G6} & Epoch     & 90       & 120       & 150      & 180      & 210    & 240    & 270    \\ \cline{2-9} 
                    & RNN(Train)       & 0.0      & 0.0      & 0.0     & 1.2e-2     & 1.0      & 1.0      & 1.0      \\ \cline{2-9} 
                    & RNN(Test)       & 0.0      & 0.0      & 0.0     & 4.7e-3     & 1.0      & 1.0      & 1.0      \\ \cline{2-9} 
                    & DFA       & 0.5      & 0.5      & 0.5     & 0.5     & 1.0      & 1.0      & 1.0      \\ \hline \hline
\multirow{4}{*}{G7} & Epoch     & 20       & 40       & 60      & 80      & 100    & 120    & 140    \\ \cline{2-9} 
                    & RNN(Train)       & 0.92      & 0.99      & 0.99     & 1.0     & 1.0      & 1.0      & 1.0      \\ \cline{2-9} 
                    & RNN(Test)       & 0.64      & 0.94      & 1.0     & 1.0     & 1.0      & 1.0      & 1.0      \\ \cline{2-9} 
                    & DFA       & 2.0e-3      & 2.0e-3      & 1.0     & 1.0     & 1.0      & 1.0      & 1.0      \\ \hline \hline
\end{tabular}
\end{table}

\begin{figure}
\centering
\begin{subfigure}{1.0\textwidth}
  \centering
  \includegraphics[width=1.0\linewidth]{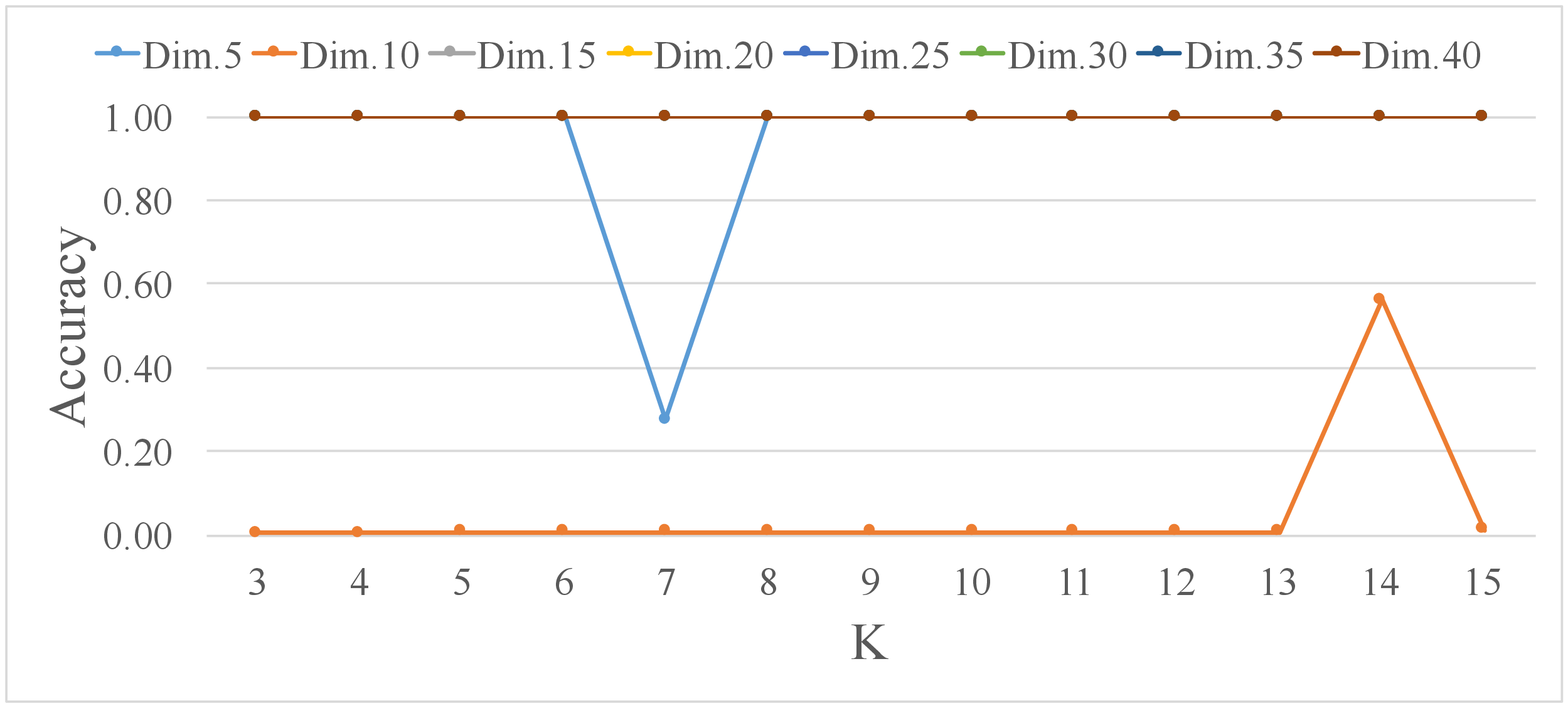}
  \caption{Influence of model capacity on DFA extraction for grammar 1.}
\end{subfigure} \\%
\begin{subfigure}{1.0\textwidth}
  \centering
  \includegraphics[width=1.0\linewidth]{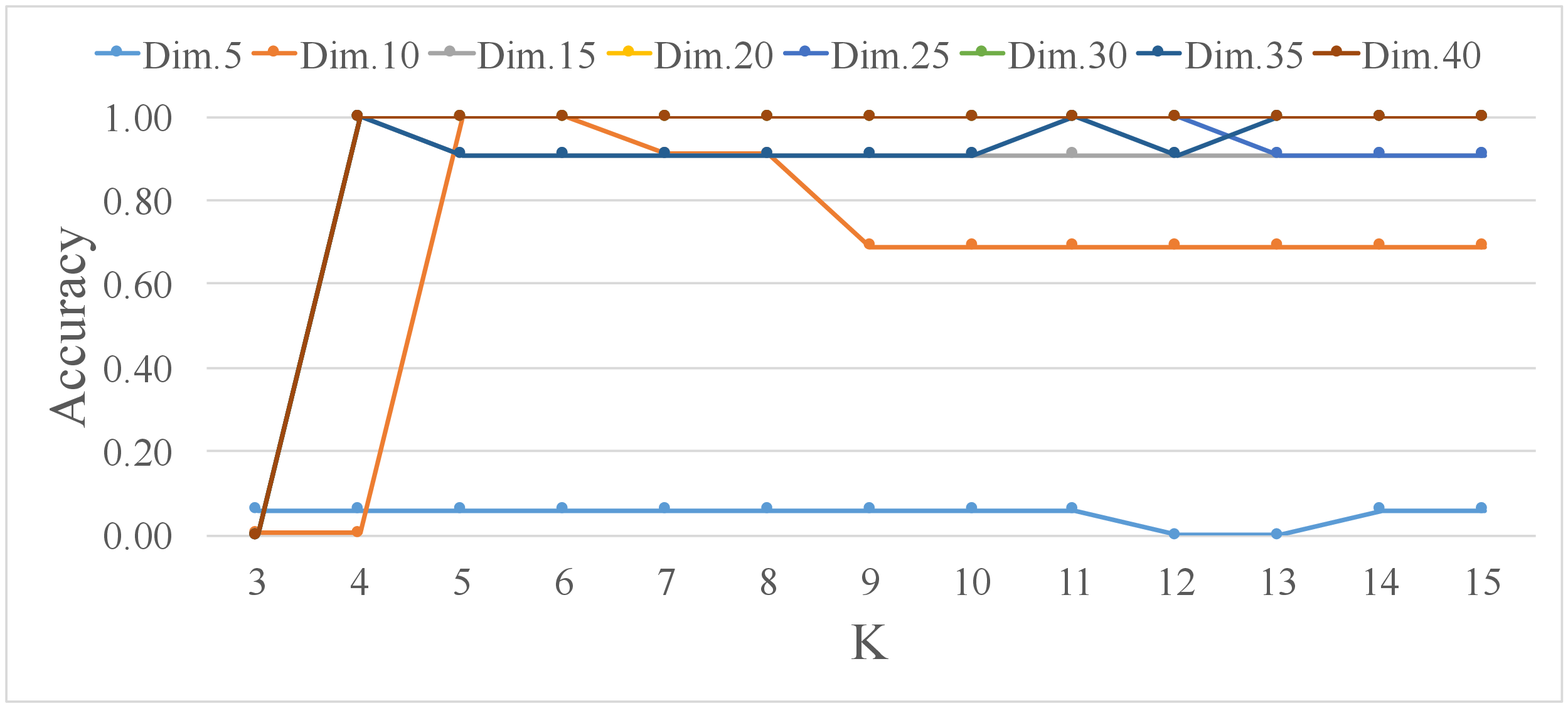}
  \caption{Influence of model capacity on DFA extraction for grammar 2.}
\end{subfigure} \\
\begin{subfigure}{1.0\textwidth}
  \centering
  \includegraphics[width=1.0\linewidth]{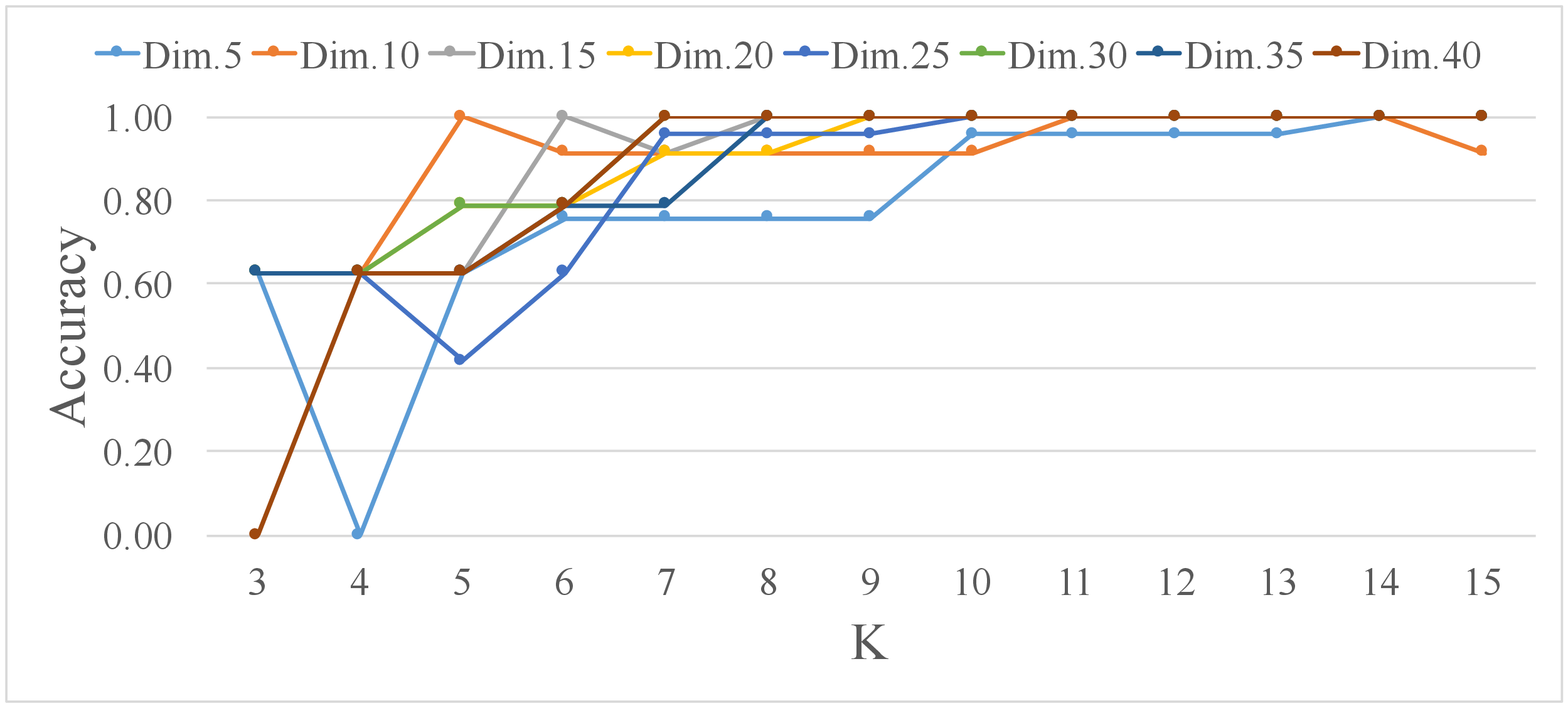}
  \caption{Influence of model capacity on DFA extraction for grammar 4.}
\end{subfigure}%
\caption{Influence of model capacity on DFA extraction for Tomita grammars.}
\label{fig:capacity_k_all}
\end{figure}

\begin{figure}\ContinuedFloat
\centering
\begin{subfigure}{1.0\textwidth}
  \centering
  \includegraphics[width=1.0\linewidth]{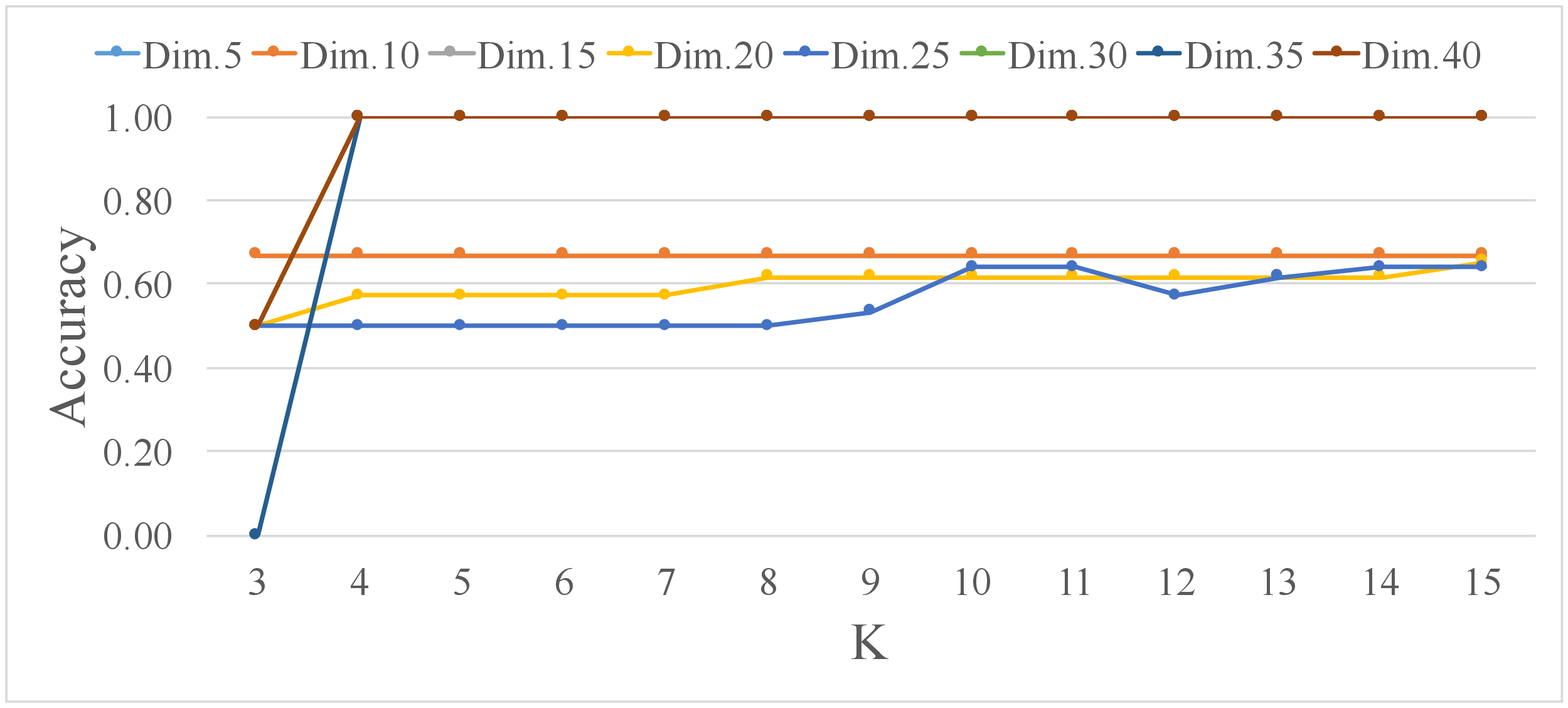}
  \caption{Influence of model capacity on DFA extraction for grammar 5.}
\end{subfigure} \\
\begin{subfigure}{1.0\textwidth}
  \centering
  \includegraphics[width=1.0\linewidth]{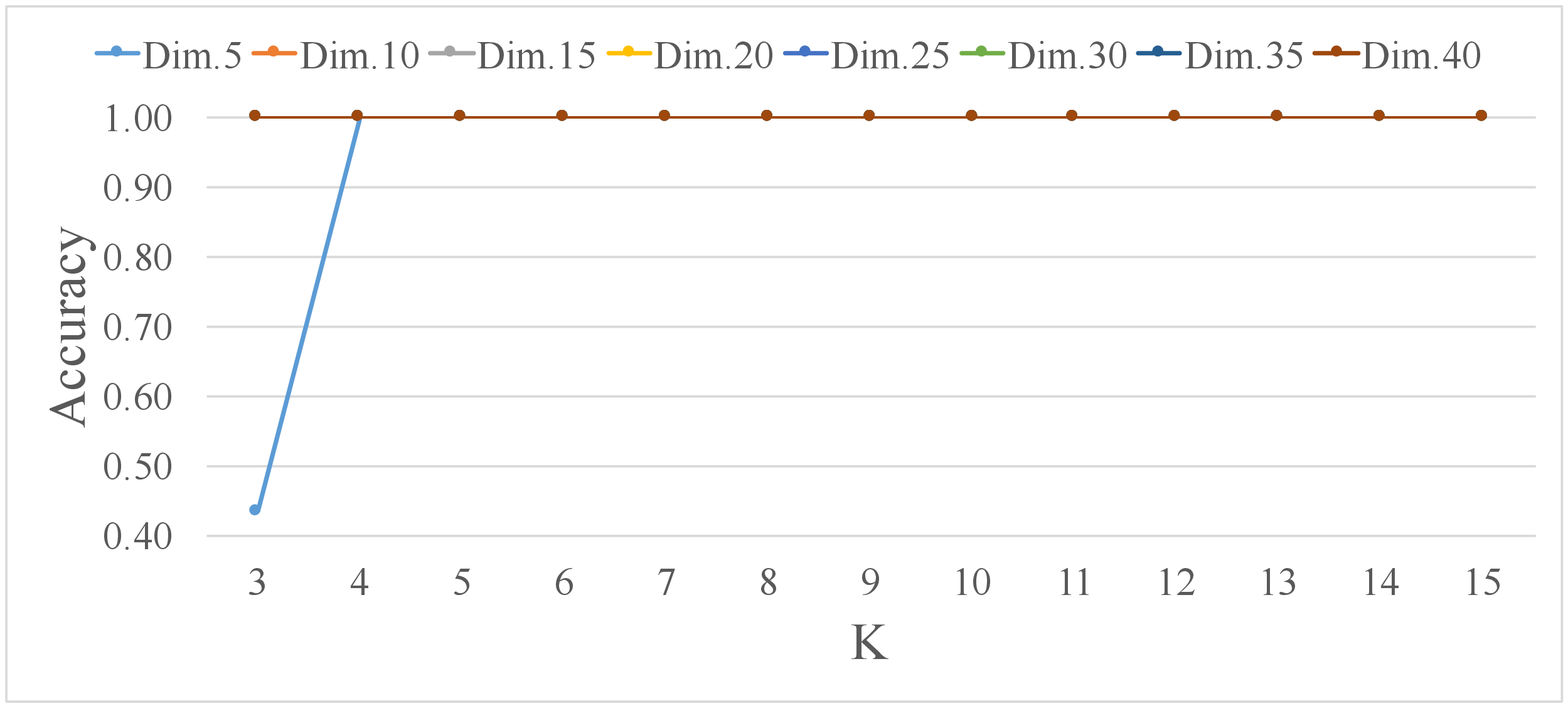}
  \caption{Influence of model capacity on DFA extraction for grammar 6.}
\end{subfigure}\\%
\begin{subfigure}{1.0\textwidth}
  \centering
  \includegraphics[width=1.0\linewidth]{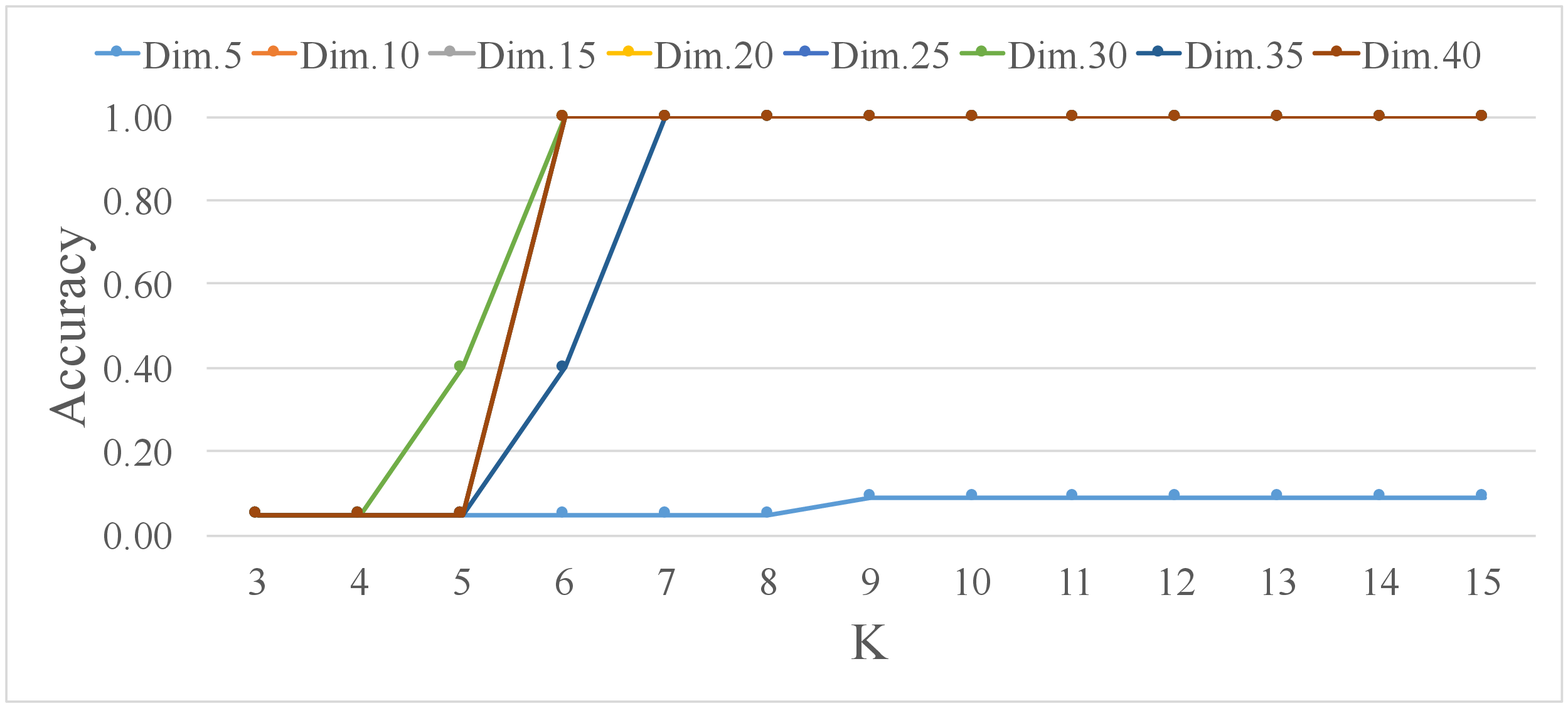}
  \caption{Influence of model capacity on DFA extraction for grammar 7.}
\end{subfigure}
\caption{Influence of model capacity on DFA extraction for Tomita grammars.}
\label{fig:capacity_k_all}
\end{figure}

\end{document}